\documentclass[sn-mathphys,Numbered]{sn-jnl}% Math and Physical Sciences Reference Style
\usepackage{graphicx}%
\usepackage{multirow}%
\usepackage{amsmath,amssymb,amsfonts}%
\usepackage{amsthm}%
\usepackage{mathrsfs}%
\usepackage[title]{appendix}%
\usepackage{xcolor}%
\usepackage{textcomp}%
\usepackage{manyfoot}%
\usepackage{booktabs}%
\usepackage{algorithm}%
\usepackage{algorithmicx}%
\usepackage{algpseudocode}%
\usepackage{listings}%

\usepackage{subfigure}
\usepackage{amsmath}
\usepackage{autobreak}
\usepackage{threeparttable}
\usepackage{bbding}

\usepackage{ragged2e} 
\usepackage{booktabs,makecell, multirow, tabularx}
\theoremstyle{thmstyleone}%
\theoremstyle{thmstyletwo}%

\theoremstyle{thmstylethree}%

\begin{document}

\title[Article Title]{Multi-proposal Collaboration and Multi-task Training for Weakly-supervised Video Moment Retrieval}


\author[1,2]{\fnm{Bolin} \sur{Zhang}}\email{onlyou@hnu.edu.cn}

\author*[1]{\fnm{Chao} \sur{Yang}}\email{yangchaoedu@hnu.edu.cn}

\author[1]{\fnm{Bin} \sur{Jiang}}\email{jiangbin@hnu.edu.cn}

\author[3]{\fnm{Takahiro} \sur{Komamizu}}\email{taka-coma@acm.org}

\author[4]{\fnm{Ichiro} \sur{Ide}}\email{ide@i.nagoya-u.ac.jp}

\affil[1]{\orgdiv{College of Computer Science and Electronic Engineering}, \orgname{Hunan University}, \orgaddress{\city{Changsha}, \postcode{410082}, \state{Hunan}, \country{China}}}

\affil[2]{\orgdiv{Faculty of Electronic Engineering and Computer Science}, \orgname{Ningbo University}, \orgaddress{\city{Ningbo}, \postcode{315211}, \state{Zhejiang}, \country{China}}}

\affil[3]{\orgdiv{Mathematical and Data Science Center}, \orgname{Nagoya University}, \orgaddress{\city{Nagoya}, \postcode{464-8601}, \state{Aichi}, \country{Japan}}}

\affil[4]{\orgdiv{Graduate School of Informatics}, \orgname{Nagoya University}, \orgaddress{\city{Nagoya}, \postcode{464-8601}, \state{Aichi}, \country{Japan}}}


\abstract{This study focuses on weakly-supervised Video Moment Retrieval (VMR), aiming to identify a moment semantically similar to the given query within an untrimmed video using only video-level correspondences, without relying on temporal annotations during training. Previous methods either aggregate predictions for all instances in the video, or indirectly address the task by proposing reconstructions for the query. However, these methods often produce low-quality temporal proposals, struggle with distinguishing misaligned moments in the same video, or lack stability due to a reliance on a single auxiliary task.
To address these limitations, we present a novel weakly-supervised method called \textbf{M}ulti-proposal \textbf{C}ollaboration and \textbf{M}ulti-task \textbf{T}raining (MCMT). Initially, we generate multiple proposals and derive corresponding learnable Gaussian masks from them. These masks are then combined to create a high-quality positive sample mask, highlighting video clips most relevant to the query. Concurrently, we classify other clips in the same video as the easy negative sample and the entire video as the hard negative sample. During training, we introduce forward and inverse masked query reconstruction tasks to impose more substantial constraints on the network, promoting more robust and stable retrieval performance. Extensive experiments on two standard benchmarks affirm the effectiveness of the proposed method in VMR.\footnote{This work was completed during the joint doctoral program at Nagoya University. }\footnote{Preprint version. The Version of Record has been published in International Journal of Machine Learning and Cybernetics. DOI: 10.1007/s13042-024-02520-w.}}




\maketitle

\section{Introduction}\label{sec1}
\begin{figure}
  \centering
  \includegraphics[width=0.92\linewidth]{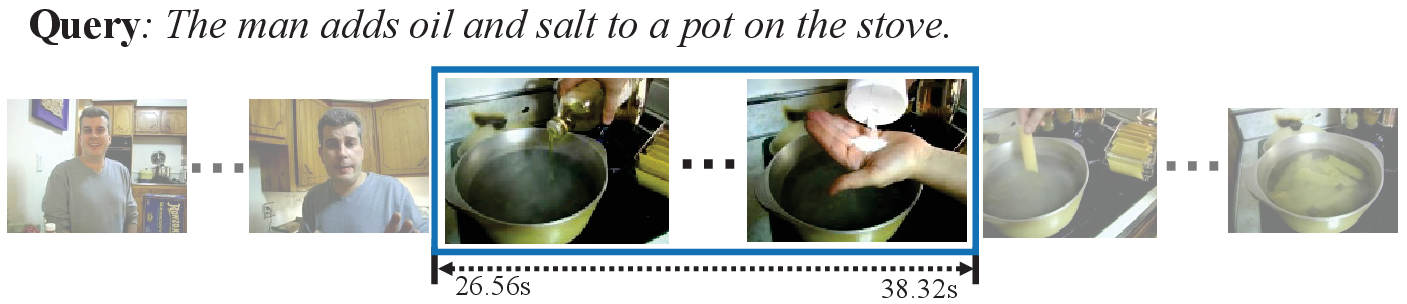} % Reduce the figure size so that it is slightly narrower than the column.
  \caption{Example of video moment retrieval given a query.}
  \label{fig1}
\end{figure}

Video Moment Retrieval (VMR) is a critical and challenging task that has numerous potential applications, such as video surveillance \cite{collins2000system,he2024video} and robot manipulation \cite{kemp2007challenges}. The objective is to automatically identify the start and end boundaries of a target moment encompassing successive clips within an untrimmed video. This moment should be semantically akin to a provided sentence query. To illustrate, when given a query like \emph{``The man adds oil and salt to a pot on the stove.''} along with a corresponding video, the goal is to retrieve the most suitable matching moment, as depicted in Figure~\ref{fig1}. Despite remarkable progress recently made in fully-supervised VMR  \cite{anne2017localizing,gao2017tall,zhang2022dual,zhang2022video,liu2018attentive}, its applicability to real-world large-scale scenarios is hindered by the labor-intensive and time-consuming process of annotating ground-truth temporal boundaries for each query sentence. Additionally, annotations are prone to inaccuracies in determining activity boundaries in videos due to their subjective nature and inconsistencies across different annotators. Consequently, the research community is increasingly drawn towards weakly-supervised settings \cite{wang2022siamese,yoon2023scanet,huang2023weakly,lv2023counterfactual} that rely only on video-level descriptions during training, making it a more practical method.

Presently, strategies employed for addressing the weakly-supervised VMR task primarily fall into two categories: multi-instance learning methods \cite{mithun2019weakly,tan2021logan,huang2021cross,yang2021local} and reconstruction-based methods \cite{duan2018weakly,lin2020weakly,chen2021towards,zheng2022weakly}. In the former methods, the input video is commonly treated as an assortment of instances, accompanied by bag-level annotations. Here, the predictions made for instances (referred to as proposal candidates) are combined to establish an overall bag-level prediction. In spite of their efficacy, these methods primarily focus on learning visual-text alignment on a video-level basis. They achieve this by maximizing the matching scores between associated queries and videos, while simultaneously attenuating the scores pertaining to unpaired instances. They also disregard the reality that distinguishing mismatched moments within the same video can present greater difficulty in the context of activity temporal localization. This is attributed to the presence of comparable backgrounds and video styles, making these instances more challenging to differentiate as negative samples.

To address this issue, CNM \cite{zheng2022weakly} generates one positive proposal and applies a Gaussian mask to extract both positive and negative samples from the same video.
However, since different video clips within the same video often share similar backgrounds, generating only a single proposal is vulnerable to redundant information from nearby video clips, especially when there is high similarity between them, which leads to inaccurate retrieval results. Moreover, existing reconstruction-based methods that indirectly address the task by producing proposals that match the query and using these proposals to reconstruct the query also have limitations. While these methods are intuitively designed and show promising results, they typically rely on reconstructing the query as a single auxiliary task, which limits the learning capability of the model and provides insufficient constraints, making the model less stable in complex scenarios.

To overcome these challenges, we present a novel \textbf{M}ulti-mask \textbf{C}ollaboration and \textbf{M}ulti-task \textbf{T}raining (MCMT) method for the weakly-supervised VMR task. Unlike the CNM method, MCMT generates multiple Gaussian mask proposals to reduce the impact of redundant information, using a mask aggregation module to improve the quality of positive samples, ensuring more accurate moment localization. MCMT dynamically generates informative positive samples and mines negative samples from within the same video by leveraging the temporal structure of events (start, climax, end) with Gaussian masks. The most relevant clips to the query are highlighted as positive samples, while unhighlighted clips are treated as easy negatives, and the entire video, containing substantial redundant information, is treated as a hard negative. By integrating multiple learnable Gaussian masks through an aggregation module, MCMT enhances the confidence in positive and negative sample mining. Additionally, MCMT introduces dual-masked reconstruction tasks, including both forward and inverse query reconstructions, to guide model training from different perspectives, providing stronger multi-angle constraints and significantly improving model stability and retrieval performance in complex scenarios.

To sum up, the main contributions of our work are:
\begin{itemize}
\item We propose a mask aggregation module to aggregate the information of multiple Gaussian masks to generate the final robust mask to mine high-quality positive, easy negative, and hard negative samples.

\item We introduce a dual-task training mechanism. In addition to the forward reconstruction masked query task, we also introduce a new task: the inverse reconstruction masked query task. During the training process, these two tasks cooperate with each other to guide the model training from different perspectives to enhance the model performance.

\item In contrast to recent methods evaluated on two benchmark datasets, the proposed MCMT method attains competitive results. As an additional contribution, we will release the codes and models to promote the development of the community\footnote{\url{https://github.com/onlyouzbl/MCMT}}.
\end{itemize}

\section{Related Work}\label{sec2}
In the weakly-supervised setting, VMR methods do not require annotations on start and end time stamps, relying only on video-query pairs. These methods aim to achieve outcomes by employing a shared multi-modal feature space or by utilizing a reconstruction-based strategy. In a general sense, current weakly-supervised VMR methods can be classified into three main groups \cite{zhang2023temporal,liu2023survey}: multi-instance learning, reconstruction-based, and other weakly-supervised methods.

\subsection{Multi-instance Learning Methods}
In the context of multi-instance learning methods, the input video is regarded as a collection of instances, with annotations assigned at the level of the entire bag. Subsequently, predictions for individual instances are aggregated to yield a comprehensive prediction for the entire bag.

Text-Guided Attention (TGA) \cite{mithun2019weakly} first solves the weakly-supervised VMR task via multi-instance learning with a text-guided attention mechanism, achieving video-text alignment by maximizing matching positive samples and minimizing negative ones. Weakly Supervised Language Localization Networks (WSLLN) \cite{gao2019wslln} measure moment-text consistency with segment selection based on text. Video-Language Alignment Network (VLANet) \cite{ma2020vlanet} refines attention through the removal of redundant proposals and the incorporation of multi-directional attention, coupled with fine-grained query representation. Weakly-Supervised Temporal Grounding (WSTG) \cite{chen2020look} proposes a two-stage model. Weakly Supervised Temporal Adjacent Network (WSTAN) \cite{wang2021weakly} achieves cross-modal semantic alignment using a self-discriminating loss along with additional branch. Latent Graph Co-Attention Network (LoGAN) \cite{tan2021logan} captures frame-word interactions through a latent graph co-attention mechanism, and Cross-Sentence Relations Mining (CRM) \cite{huang2021cross} examines inter-sentence connections to refine video moment proposal selections. Despite their effectiveness, such methods focus on video-text alignment at the video level by maximizing matching scores, while overlooking the difficulty of distinguishing mismatched moments within the same video because of similar backgrounds and video styles. Inspired by CNM \cite{zheng2022weakly}, we use the Gaussian mask to fit the temporal structure of the event (start, climax, end). To generate a more robust mask, we first generate multiple learnable Gaussian masks and then integrate these masks together by the proposed aggregation module to mine high-quality positive and negative samples with higher confidence in the same video.

\begin{figure*}[ht]
  \centering
  \includegraphics[width=1.0\linewidth]{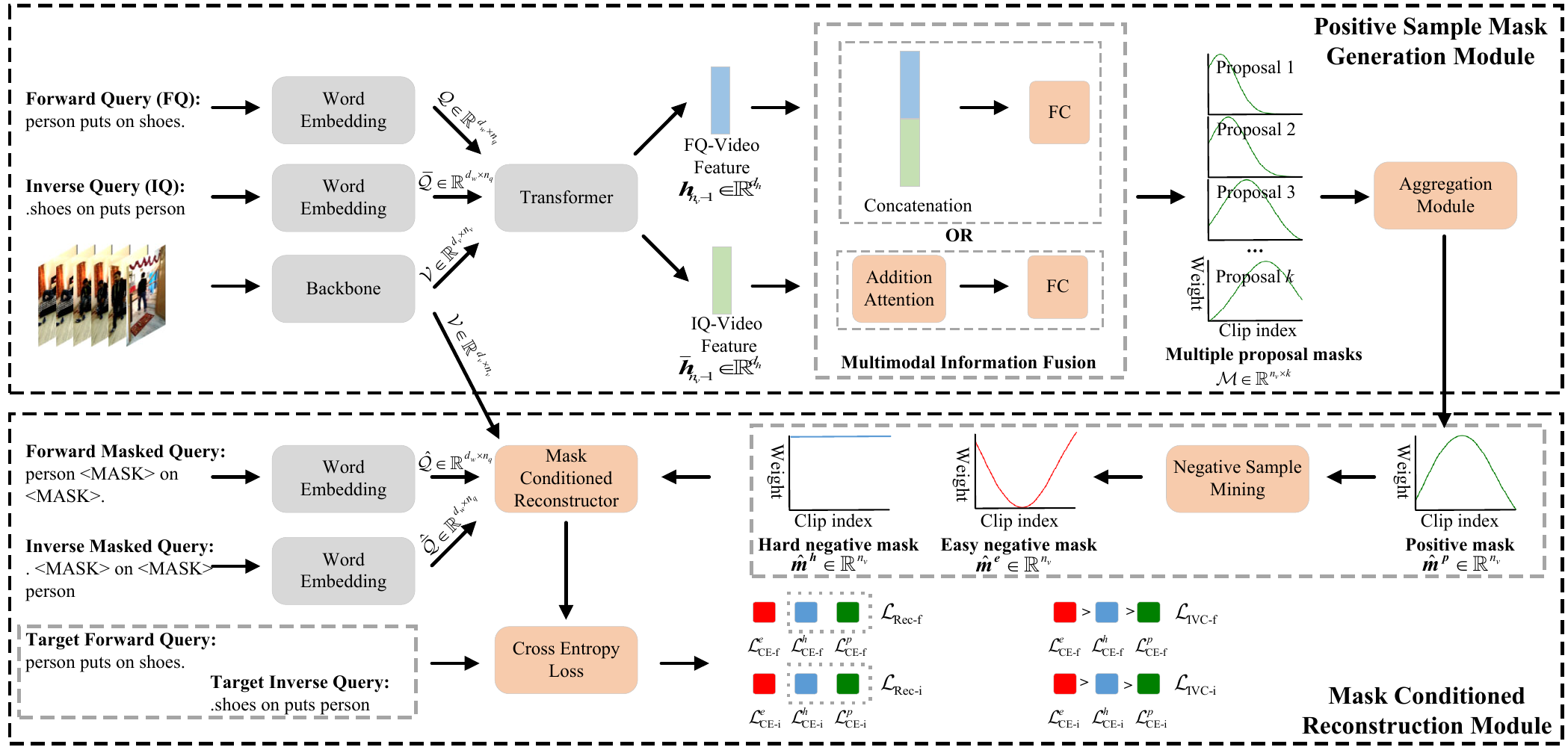} % Reduce the figure size so that it is slightly narrower than the column.
  \caption{Overview of the proposed MCMT model. It mainly includes a positive sample mask generation module (Section 3.1), and a mask conditioned reconstruction module (Section 3.2). The former fuses information from queries and video modalities and generates multiple candidate proposals, followed by using the proposed aggregation module to generate the final high-quality positive sample mask. The latter first mines high-quality hard negative and easy negative masks using the positive mask and then reconstructs the forward and inverse masked queries to measure the similarity of positive and negative samples and queries. Forward Reconstruction loss $\mathcal{L}_\mathrm{Rec\text{-}f}$, Inverse Reconstruction loss $\mathcal{L}_\mathrm{Rec\text{-}i}$, forward Intra-Video Contrastive loss $\mathcal{L}_\mathrm{IVC\text{-}f}$ and inverse Intra-Video Contrastive loss $\mathcal{L}_\mathrm{IVC\text{-}i}$ jointly guide the model training.}
  \label{fig2}
\end{figure*}

\subsection{Reconstruction-based Methods}
Reconstruction-based methods indirectly tackle the task by generating proposals that match the query from the video input. These proposals are subsequently employed to reconstruct the query, with the intermediate proposals serving as retrieval results.

Weakly Supervised Dense Event Captioning (WS-DEC) \cite{duan2018weakly} proposes weakly supervised dense event captioning by decomposing the problem into event captioning and sentence localization. Semantic Completion Network (SCN) \cite{lin2020weakly} selects top-$k$ video proposals, computes rewards based on reconstruction loss, and refines proposal generation. Multi-level Attentional Reconstruction Network (MARN) \cite{song2020weakly} learns a language-driven attention map by using intra-proposal and inter-proposal interaction. Event Captioner and Sentence Localizer (EC-SL) \cite{chen2021towards} introduces an Induced Set Attention Block to establish a connection between the event captioner and the sentence localizer. CNM \cite{zheng2022weakly} generates the positive sample by using a learnable Gaussian mask that highlights video clips most relevant to the query, and treats other clips and the whole video as easy and hard negative samples, respectively. The Intra-Video Contrastive (IVC) loss is used to train the network and improve the discriminative ability of the positive and negative samples. These methods achieve some good results but may lack sufficient constraints by only reconstructing the masked query as a single auxiliary task during training, resulting in less stable outcomes. We propose a dual-task training mechanism that involves both forward and inverse reconstruction masked query tasks to guide model training from different perspectives and enhance the model performance.

\subsection{Other Weakly-supervised Methods}
In addition to the above two method categories, Regularized Two-Branch Proposal Networks (RTBPN) \cite{zhang2020regularized} consider both inter- and intra-sample confrontments to address the difficulty of distinguishing the target moment from plausible negative moments that current standard multi-instance learning methods face due to their focus on inter-sample confrontment only. Pseudo-Supervised Video Localization (PSVL) \cite{nam2021zero} introduces a novel zero-shot natural language video localization task that requires no paired annotation cost and proposes a pseudo-supervised method, that generates pseudo-supervision for training. Without using any textual annotations of videos, Unpaired Video Moment Retrieval (U-VMR) \cite{gao2021learning} employs a pre-trained image-sentence embedding space and utilizes established visual concept detectors for VMR.

\begin{table}[t]
\centering
\caption{Acronyms and Symbols}
\begin{tabular}{|c|l|}
\hline
\textbf{Acronym/Symbol} & \textbf{Definition}                                \\ \hline
$\mathcal{W}$           & Sentence query                                     \\ \hline
$\mathcal{F}$           & Untrimmed video                                    \\ \hline
$n_{q}$                 & Number of words in the query                       \\ \hline
$w_{i}$                 & Word in the sentence                               \\ \hline
$\boldsymbol{q}_{i}$    & Word feature vector                                \\ \hline
$\mathcal{Q}$           & Word embeddings                                    \\ \hline
$f_{i}$                 & Video frame                                        \\ \hline
$n_f$                   & Number of frames in the video                      \\ \hline
$v_{i}$                 & Video clip                                         \\ \hline
$n_{v}$                 & Number of video clips                              \\ \hline
$\boldsymbol{v}_{i}$    & Video clip feature vector                          \\ \hline
$\mathcal{V}$           & Visual features                                    \\ \hline
$\mathcal{H}$           & Fused forward query-video features                 \\ \hline
$\mathcal{\bar{H}}$     & Fused inverse query-video features                 \\ \hline
$\textit{E}(\cdot)$     & Transformer encoder function                       \\ \hline
$\textit{D}(\cdot)$     & Transformer decoder function                       \\ \hline
$\boldsymbol{c}$        & Center of $k$ proposal candidates                  \\ \hline
$\boldsymbol{w}$        & Width of $k$ proposal candidates                   \\ \hline
$\mathcal{M}$           & Matrix containing $k$ Gaussian masks               \\ \hline
$\boldsymbol{\hat{m}^p}$  & Positive sample mask                             \\ \hline
$\boldsymbol{\hat{m}^e}$  & Easy negative mask                               \\ \hline
$\boldsymbol{\hat{m}^h}$  & Hard negative mask                               \\ \hline
$\mathcal{\hat{Q}}$ & Masked forward query embedding                         \\ \hline
$\mathcal{\hat{\bar{Q}}}$ & Masked inverse query embedding                   \\ \hline
$\mathcal{\hat{H}}$ & Forward cross-modal semantic representation            \\ \hline
$\mathcal{\hat{\bar{H}}}$ & Inverse cross-modal semantic representation      \\ \hline
$\mathcal{L}_\mathrm{CE\text{-}f}^p$ & Forward cross-entropy loss of positive mask \\ \hline
$\mathcal{L}_\mathrm{CE\text{-}i}^p$ & Inverse cross-entropy loss of positive mask \\ \hline
$\mathcal{L}_\mathrm{Rec}$               & Reconstruction loss                                \\ \hline
$\mathcal{L}_\mathrm{IVC}$               & Intra-Video Contrastive loss                       \\ \hline
\end{tabular}
\label{tabacr}
\end{table}

\section{Proposed Method}
In this section, we detail the proposed Multi-proposal Collaboration and Multi-task Training method (MCMT) from two components, positive sample mask generation module and mask conditioned reconstruction module, as illustrated in Figure \ref{fig2}. In addition,  the acronym and symbol table is as shown in Table~\ref{tabacr}, providing clear definitions for each variable and acronym to enhance readability.

\subsection{Positive Sample Mask Generation Module}
Based on the content of the video and the query, this module fully incorporates information from both modalities to generate a high-quality positive mask. To characterize the inherent temporal structure of the event (start, climax, end), following CNM \cite{zheng2022weakly}, we use Gaussian masks as the proposals. 
Considering that CNM only generates a single Gaussian mask as the positive proposal, it becomes susceptible to redundant information from nearby video clips with similar backgrounds. This susceptibility leads to less stable retrieval performance. To address this limitation, the proposed method predicts multiple Gaussian masks simultaneously. Subsequently, the information from these masks is aggregated to generate a final high-quality positive mask. This final mask serves to highlight video clips that are semantically similar to the query.
Due to the operation of aggregating multiple masks, the proposed method can detect video clips that are semantically similar to the query in a more efficient and stable way, resulting in more accurate localization.

\subsubsection{Feature Extraction}
Let $\mathcal{W}$ denote a sentence query, and $\mathcal{F}$ denote a corresponding untrimmed video. The query $\mathcal{W}=\left[w_{i}\right]_{i=0}^{n_{q}-1}$ contains $n_{q}$ words, with $w_{i}$ representing a word in the sentence.

For words in $\mathcal{W}$, following previous methods~\cite{lin2020weakly,ma2020vlanet,yang2021local,chen2021towards,zheng2022weakly}, we also apply GloVe~\cite{pennington2014glove} to encode word embeddings as $\mathcal{Q}=\left[\boldsymbol{q}_{i}\right]_{i=0}^{n_{q}-1} \in \mathbb{R}^{d_{w} \times n_{q}}$, where $\boldsymbol{q}_{i}$ represents the word feature vector, $d_{w}$ is the word feature dimension. We have chosen GloVe as the text feature extraction tool primarily because of its ability to capture semantic relationships and similarities between words. By pre-training on large-scale corpora in an unsupervised manner, GloVe effectively captures the global semantics of words. Compared to randomly initialized word embeddings, GloVe provides more semantically representative word vectors, which is especially beneficial in weakly-supervised video moment retrieval tasks, where it enhances the alignment of semantic information between text queries and video moments.

The video $\mathcal{F}=\left[f_{i}\right]_{i=0}^{n_{f}-1}$ contains a total sequence of $n_{f}$ frames, with each frame denoted as $f_{i}$. Each video $\mathcal{F}$ is divided into clip units, each consisting of $T$ frames. Subsequently, $n_v$ video clips are sampled at a constant interval, resulting in $\mathcal{F}=\left[v_{i}\right]_{i=0}^{n_{v}-1}$, which are then input to a pre-trained feature extractor \cite{radford2021learning,carreira2017quo} for feature extraction. The extracted features are forwarded to a fully connected layer with $d_v$ channels to acquire a compact feature representation. In conclusion, the visual features $\mathcal{V}=\left[\boldsymbol{v}_{i}\right]_{i=0}^{n_{v}-1} \in \mathbb{R}^{d_{v} \times n_{v}}$ are generated, where $d_{v}$ denotes the dimension of the visual features. VMR is posed as an objective to retrieve the optimal corresponding video moment $\boldsymbol{m}^{*}=\left\{\boldsymbol{v}_{i} \mid i=i^{s}, \cdots, i^{e}\right\}$, which initiates at clip index $i^s$ and concludes at $i^e$. This video moment is selected based on its capacity to convey the same semantic content as the provided sentence query $\mathcal{W}$.

\subsubsection{Positive Mask Generation}
Following CNM, we use Gaussian masks to characterize the inherent temporal structure (start, climax, end). We use transformer \cite{vaswani2017attention} to perform cross-modal interactions and obtain the fused forward query-video features $\mathcal{H}$ and inverse query-video features $\mathcal{\bar{H}}$ by:
\begin{equation}
\begin{aligned}
\mathcal{H}=\left[\boldsymbol{h}_{i}\right]_{i=0}^{n_{v}-1}=\textit{D}(\mathcal{V}, \textit{E}(\mathcal{Q})) \in \mathbb{R}^{d_h \times n_v} , \\
\mathcal{\bar{H}}=\left[\boldsymbol{\bar{h}}_{j}\right]_{j=0}^{n_{v}-1}=\textit{D}(\mathcal{V}, \textit{E}(\mathcal{\bar{Q}})) \in \mathbb{R}^{d_h \times n_v} ,
\end{aligned}
\end{equation}
where $\textit{E}(\cdot)$ is the transformer encoder, $\textit{D}(\cdot)$ is the transformer decoder, $d_h$ is the hidden feature dimension. Specifically, since $\boldsymbol{h}_{n_v-1} \in \mathbb{R}^{d_h}$ and $\boldsymbol{\bar{h}}_{n_v-1} \in \mathbb{R}^{d_h}$ fuse all clip and word features, we design two mechanisms to perform cross-modal feature fusion: vector concatenation and additive attention \cite{bahdanau2015neural}.

\textbf{Vector Concatenation Mechanism.} 
We use vector concatenation ($||$) followed by a Fully Connected (FC) layer to fuse information of all modalities and finally predict the center $\boldsymbol{c}$ and width $\boldsymbol{w}$ of $k$ proposal candidates by the Sigmoid function:
\begin{equation}
\begin{aligned}
(\boldsymbol{c, w})=\operatorname{Sigmoid}\left(\left(\boldsymbol{h}_{n_v-1} || \boldsymbol{\bar{h}}_{n_v-1}\right) \cdot \boldsymbol{W}_p \right) & \in \mathbb{R}^{2k}, \\
\boldsymbol{c}\in \mathbb{R}^{k}, \boldsymbol{w} & \in \mathbb{R}^{k}, 
\end{aligned}
\label{equ2}
 \end{equation}
where $\boldsymbol{W}_p \in \mathbb{R}^{2d_h \times 2k}$ is the weight for the FC prediction layer.

\textbf{Additive Attention Mechanism.} We apply additive attention mechanism \cite{bahdanau2015neural} to compute the attention scores of $\boldsymbol{h}_{n_v-1} \in \mathbb{R}^{d_h}$ and $\boldsymbol{\bar{h}}_{n_v-1} \in \mathbb{R}^{d_h}$. The scores computed are utilized to aggregate the information of $\widetilde{\boldsymbol{h}}=\left[ \boldsymbol{h}_{n_v-1},\boldsymbol{\bar{h}}_{n_v-1}\right]$ to compute the modularized fused vector $\boldsymbol{\hat{h}}$:
\begin{equation}
\boldsymbol{\beta}=\operatorname{Softmax}\left(\boldsymbol{W}_h \cdot \widetilde{\boldsymbol{h}} \right) \in \mathbb{R}^{2}, \quad \boldsymbol{\hat{h}}=\sum_{i=0}^{1} \boldsymbol{\beta}_i \times \widetilde{\boldsymbol{h}} \in \mathbb{R}^{d_h},
\end{equation}
where $\boldsymbol{W}_h \in \mathbb{R}^{1 \times d_h}$ is the weight for the modularized feed-forward layer, $\boldsymbol{\beta}$ represents the attention weights computed through the additive attention mechanism. These weights are derived by applying a softmax function to the vector 
$\boldsymbol{\hat{h}}$, which consists of the fused forward and inverse query-video features. The attention weights  $\boldsymbol{\beta}$ are used to determine the contribution of each modality to the final fused feature $\boldsymbol{\hat{h}}$, where the sum of all attention weights equals 1, ensuring a weighted balance between different components. Then we also use an FC layer and predict the center $\boldsymbol{c}$ and width $\boldsymbol{w}$ of $k$ proposal candidates by the Sigmoid function:
\begin{equation}
(\boldsymbol{c, w})=\operatorname{Sigmoid}(\boldsymbol{\hat{h}} \cdot \boldsymbol{W}_p) \in \mathbb{R}^{2k}, \boldsymbol{c}\in \mathbb{R}^{k}, \boldsymbol{w} \in \mathbb{R}^{k},
\label{eq4}
\end{equation}
where $\boldsymbol{W}_p \in \mathbb{R}^{d_h \times 2k}$ is the weight for the FC prediction layer.

We explore these two separate multi-modal fusion strategies, each applied independently to compute the center and width of the proposal candidates. Based on these predicted $k$ proposal candidates, we generate the corresponding $k$ Gaussian masks:
\begin{equation}
\mathcal{M}= \\ 
\left[\left[\boldsymbol{m}_{ij}\right]_{j=0}^{n_v-1}\right]_{i=0}^{k-1}=\exp \left(-\frac{\alpha((j+1) / {n_v}-\boldsymbol{c}_i)^2}{\boldsymbol{w}_i^2}\right)  \in \mathbb{R}^{n_v \times k},
\end{equation}
where $\boldsymbol{m}_{ij}$ signifies the weight attributed to the $j$-th video clip within the $i$-th Gaussian mask, while $\alpha$ is a hyperparameter governing the variance of the Gaussian function.

\subsubsection{Mask Aggregation Module}
To generate a high-quality positive sample mask, we propose a mask aggregation module to aggregate the information of the generated $k$ Gaussian masks. Specifically, we also apply an additive attention mechanism \cite{bahdanau2015neural} to calculate the attention scores for each Gaussian mask. These computed scores are employed to consolidate the information from $\mathcal{M}=\left[ m_0,m_1, \cdots, m_{k-1} \right]$ and compute the modularized mask vector $\boldsymbol{\hat{m}^p}$, serving as a positive mask:
\begin{equation}
\left.\boldsymbol{\beta}=\operatorname{Softmax}\left(\boldsymbol{W}_m \cdot \mathcal{M}\right)\right) \in \mathbb{R}^{k}, \quad \boldsymbol{\hat{m}^p}=\sum_{i=0}^{n_v-1} \boldsymbol{\beta}_i \times \boldsymbol{m}_i \in \mathbb{R}^{n_v},
\label{equ6}
\end{equation}
where $\boldsymbol{W}_m \in \mathbb{R}^{1 \times n_v}$ is the weight for the modularized mask feed-forward layer.

\subsection{Mask Conditioned Reconstruction Module}
In order to empower the introduced model with the capability to differentiate intricate scenes effectively, we initially mine the negative mask from within the same video as the positive mask. Then, mask conditioned reconstruction module reconstructs the forward and inverse queries based on an arbitrary mask, and the reconstructed results are used as a measure of the semantic similarity between clips highlighted by positive/negative masks and the query. In addition to reconstruction query losses $\mathcal{L}_\mathrm{Rec\text{-}f}$ and $\mathcal{L}_\mathrm{Rec\text{-}i}$, following CNM, we also introduce the Intra-Video Contrastive (IVC) losses $\mathcal{L}_\mathrm{IVC\text{-}f}$ and $\mathcal{L}_\mathrm{IVC\text{-}i}$ to optimize the generated masks in an end-to-end manner.

\subsubsection{Negative Sample Mining}
Unlike CNM \cite{zheng2022weakly} that directly uses the generated single Gaussian mask as a positive mask, the proposed method aggregates the information of multiple predicted Gaussian masks to generate a robust positive mask, and mine negative masks based on this high-quality positive mask.

To enable the proposed model to discriminate highly complex scenes, following CNM, instead of simply using other mismatched videos as negative samples, we mine two types of negative samples within the same video: easy negative and hard negative.

Specifically, we regard the video clips suppressed by the positive sample mask $\boldsymbol{\hat{m}^p}$ as an easy negative sample $\boldsymbol{\hat{m}^e}$ as:
\begin{equation}
\boldsymbol{\hat{m}^e}=1-\boldsymbol{\hat{m}^p} \in \mathbb{R}^{n_v}.
\end{equation}
In addition, since the entire video contains a lot of redundant information that is not relevant to the query, we treat the entire video as a hard negative sample $\boldsymbol{\hat{m}^h}$, denoted as:
\begin{equation}
\boldsymbol{\hat{m}^h}=[1,1, \ldots, 1] \in \mathbb{R}^{n_v}.
\end{equation}
By discriminating between positive and hard/easy negative samples, we can improve the performance of the proposed model in highly complex scenes.

Given that the entire video encompasses a considerable amount of redundant information alongside the ground-truth clips, and an easy negative sample does not encompass any correct clips. In this context, the semantic similarity between these samples and the query is expected to fulfill:
\begin{equation}
R\left(\boldsymbol{\hat{m}^p}, \mathcal{Q}\right)>R\left(\boldsymbol{\hat{m}^h}, \mathcal{Q}\right)>R\left(\boldsymbol{\hat{m}^e}, \mathcal{Q}\right),
\label{equ9}
\end{equation}
where $R(\cdot)$ is a function that computes the semantic similarity between the query and the video clip represented by mask $\boldsymbol{\hat{m}}$.

\subsubsection{Mask Conditioned Semantic Completion}
We assume that given a forward or a inverse masked query, the sample most relevant to the original query should be able to reconstruct the entire query better. To contrast positive and negative samples, following SCN \cite{lin2020weakly} and CNM \cite{zheng2022weakly}, we introduce the masked query reconstruction task to measure the semantic similarity of queries and positive and negative samples. In addition, we also introduce the inverse masked query reconstruction task to measure the similarity between inverse queries and positive and negative samples. These two tasks collaborate with each other to better compare positive and negative samples, thus achieving better performance.

First, we use a special symbol to randomly mask 1/3 of the words of the original forward and inverse queries, where nouns, verbs, and adjectives have a higher probability of being replaced. The purpose is to balance learning efficiency and computational burden.
If too few words are masked, the model may not face enough semantic inference challenges, whereas masking too many words might lead to excessive information loss, hindering the training process. The 1/3 masking ratio provides sufficient inference space for the model while retaining enough contextual information, ensuring that the model can effectively learn the semantic relationships between words during the training process.

Then, we use GloVe \cite{pennington2014glove} to encode the masked forward and inverse queries and pass an FC layer to obtain the masked forward and inverse query embedding $\mathcal{\hat{Q}} \in \mathbb{R}^{d_{w} \times n_{q}}$ and $\mathcal{\hat{\bar{Q}}} \in \mathbb{R}^{d_{w} \times n_{q}}$, respectively. To prevent leakage of video features located outside the mask and to make the entire module differentiable to the mask, we use the mask conditioned transformer proposed by CNM \cite{zheng2022weakly} to reconstruct the original forward and inverse queries. Specifically, it multiplies the positive mask $\boldsymbol{\hat{m}^p}$ by the attention map before aggregating contextual information, and finally, we obtain the forward and inverse cross-modal semantic representations $\mathcal{\hat{H}}$ and $\mathcal{\hat{\bar{H}}}$:
\begin{equation}
\begin{aligned}
\mathcal{\hat{H}}=D_m\left(\mathcal{\hat{Q}}, E_m\left(\mathcal{V}, \boldsymbol{\hat{m}^p}\right), \boldsymbol{\hat{m}^p}\right) \in \mathbb{R}^{d \times n_q}, \\
\mathcal{\hat{\bar{H}}}=D_m\left(\mathcal{\hat{\bar{Q}}}, E_m\left(\mathcal{V}, \boldsymbol{\hat{m}^p}\right), \boldsymbol{\hat{m}^p}\right) \in \mathbb{R}^{d \times n_q},
\end{aligned}
\end{equation}
where $d$ is the cross-modal feature dimension. Then, $\mathcal{\hat{H}}$ and $\mathcal{\hat{\bar{H}}}$ are fed into the FC prediction layer, and finally, given the first $i-1$ words, we obtain the probability distribution of the $i$-th word:
\begin{equation}
\begin{aligned}
\mathcal{P}_\mathcal{\hat{H}}\left(\widetilde{\boldsymbol{q}}_{i} \mid \mathcal{V},\mathcal{\hat{Q}}_{0:i-1}\right)=\operatorname{Softmax}\left(\mathrm{FC}\left(\mathcal{\hat{H}}\right)\right) \in \mathbb{R}^{N_q \times n_q}, \\
\mathcal{P}_\mathcal{\hat{\bar{H}}}\left(\widetilde{\boldsymbol{\bar{q}}}_{i} \mid \mathcal{V},\mathcal{\hat{\bar{Q}}}_{0:i-1}\right)=\operatorname{Softmax}\left(\mathrm{FC}\left(\mathcal{\hat{\bar{H}}}\right)\right) \in \mathbb{R}^{N_q \times n_q},
\end{aligned}
\end{equation}
where $N_q$ is the vocabulary size. 

Finally, we can calculate the forward and inverse cross-entropy losses $\mathcal{L}_\mathrm{CE\text{-}f}^p$ and $\mathcal{L}_\mathrm{CE\text{-}i}^p$ of the positive sample by:
\begin{equation}
\begin{aligned}
\mathcal{L}_\mathrm{CE\text{-}f}^p=-\sum_{i=0}^{n_q-1} \log \mathcal{P}_\mathcal{\hat{H}}\left(\widetilde{\boldsymbol{q}}_{i} \mid \mathcal{V},\mathcal{\hat{Q}}_{0:i-1}\right), \\
\mathcal{L}_\mathrm{CE\text{-}i}^p=-\sum_{i=0}^{n_q-1} \log \mathcal{P}_\mathcal{\hat{\bar{H}}}\left(\widetilde{\boldsymbol{\bar{q}}}_{i} \mid \mathcal{V},\mathcal{\hat{\bar{Q}}}_{0:i-1}\right).
\end{aligned}
\end{equation}
Similarly, by replacing $\boldsymbol{\hat{m}^p}$ with $\boldsymbol{\hat{m}^e}$ and $\boldsymbol{\hat{m}^h}$, we can compute $\mathcal{L}_\mathrm{CE\text{-}f}^e$, $\mathcal{L}_\mathrm{CE\text{-}i}^e$, $\mathcal{L}_\mathrm{CE\text{-}f}^h$, and $\mathcal{L}_\mathrm{CE\text{-}i}^h$. We discard $\mathcal{L}_\mathrm{CE\text{-}f}^e$ and $\mathcal{L}_\mathrm{CE\text{-}i}^e$ 
because an easy sample does not contain query-related clips to reconstruct the original forward and inverse queries, so the cross-entropy loss of positive and hard negative samples (the entire video) is retained to participate in the optimization of the mask. The reconstruction loss is:
\begin{equation}
\mathcal{L}_\mathrm{Rec}=\mathcal{L}_\mathrm{CE\text{-}f}^p+\mathcal{L}_\mathrm{CE\text{-}i}^p+\mathcal{L}_\mathrm{CE\text{-}f}^h+\mathcal{L}_\mathrm{CE\text{-}i}^h.
\end{equation}
As shown in Eq.\ref{equ9}, the semantic similarity of positive, hard negative, and easy negative samples to the query needs to satisfy a relationship from large to small. Following CNM, we use the IVC loss of two tasks to compare positive and negative samples:
\begin{equation}
\begin{aligned}
\mathcal{L}_\mathrm{IVC\text{-}f}= & \max \left(\mathcal{L}_\mathrm{CE\text{-}f}^p-\mathcal{L}_\mathrm{CE\text{-}f}^h+\beta_1, 0\right)+ \\
& \max \left(\mathcal{L}_\mathrm{CE\text{-}f}^p-\mathcal{L}_\mathrm{CE\text{-}f}^e+\beta_2, 0\right),
\end{aligned}
\end{equation}
\begin{equation}
\begin{aligned}
\mathcal{L}_\mathrm{IVC\text{-}i}= & \max \left(\mathcal{L}_\mathrm{CE\text{-}i}^p-\mathcal{L}_\mathrm{CE\text{-}i}^h+\beta_3, 0\right)+ \\
& \max \left(\mathcal{L}_\mathrm{CE\text{-}i}^p-\mathcal{L}_\mathrm{CE\text{-}i}^e+\beta_4, 0\right),
\end{aligned}  
\end{equation}
where $\beta_1$ and $\beta_2$ are hyperparameters satisfying $\beta_1 < \beta_2$. The condition imposed by $\mathcal{L}_\mathrm{IVC-f}$ mandates that the loss associated with the positive sample must be a minimum of $\beta_1$ lower than that of the hard negative sample, and at least $\beta_2$ lower than the loss of the easy negative sample. Similarly, $\beta_3 < \beta_4$. The final IVC loss $\mathcal{L}_\mathrm{IVC}$ is:
\begin{equation}
\mathcal{L}_\mathrm{IVC}=\mathcal{L}_\mathrm{IVC\text{-}f}+\mathcal{L}_\mathrm{IVC\text{-}i}.
\end{equation}
\subsection{Model Training and Inference}
\subsubsection{Training}
The reconstruction loss  $\mathcal{L}_\mathrm{Rec}$ optimizes the mask conditioned reconstruction module to predict the query related to the given mask. The IVC loss $\mathcal{L}_\mathrm{IVC}$ enhances the optimization of the mask generation module, aiming to increase the distinctiveness between positive and negative samples. Following CNM \cite{zheng2022weakly}, the proposed method sequentially updates the reconstructor with $\mathcal{L}_\mathrm{Rec}$ while freezing the mask generator, followed by updating the mask generator with $\mathcal{L}_\mathrm{IVC}$ while freezing the reconstructor.

\subsubsection{Inference}
First, the center $\boldsymbol{c}$ and width $\boldsymbol{w}$ in Eq.\ref{equ2} are calculated for the predicted $k$ proposals, as well as the attention scores $\boldsymbol{\beta}$ in Eq.\ref{equ6} which indicate the reliability of each proposal's contribution to the positive mask $\boldsymbol{\hat{m}^p}$. Next, the proposals are sorted by their corresponding scores in descending order to obtain the top-$k$ predictions. Specifically, to select the top-1 prediction from the $k$ proposals, we adopt a vote-based strategy \cite{zhou2009ensemble,2zheng2022weakly} and use the $k$ proposals to vote with each other and select the one with the highest number of votes as the final top-1 prediction, where the IoU is calculated between each proposal and the remaining $k-1$ proposals to determine the number of votes each proposal gains. Ultimately, the temporal boundaries $(\boldsymbol{s}, \boldsymbol{e})$ of the top-$k$ propsals can be obtained by:
\begin{equation}
\begin{aligned}
& \boldsymbol{s}=\max (\boldsymbol{c}-\boldsymbol{w} / 2,0) \times  d ,\\
& \boldsymbol{e}=\min (\boldsymbol{c}+\boldsymbol{w} / 2,1) \times  d ,
\end{aligned}
\end{equation}
where $d$ denotes the duration of the corresponding untrimmed video.

\begin{table}
%\small
% \footnotesize
\centering
\scriptsize
\caption{Summary of each dataset.}
\begin{tabular}{ccccc}
\toprule
Dataset      & \#Videos & \#Anno. (train/val/test)  & $L_\mathrm{vid}$ &             $L_\mathrm{mom}$ \\  
\midrule 
Charades & \multicolumn{1}{r}{6,672}    & 12,408 / --- / 3,720   &  \multicolumn{1}{r}{30.59s} & \multicolumn{1}{r}{8.22s}\\ 
ActivityNet & \multicolumn{1}{r}{19,207}   & 37,417 / 17,505 / 17,031        & \multicolumn{1}{r}{117.61s} & \multicolumn{1}{r}{36.18s} \\ 
\bottomrule
\end{tabular}                   
\footnotetext[1]{\#Anno.: Number of query-moment pairs in different datasets.}   
\footnotetext[2]{$L_\mathrm{vid}$ and $L_\mathrm{mom}$: Average lengths of videos and moments, respectively.}       
\label{tab1}          
\end{table}

\begin{table}
%\small
% \footnotesize
\centering
\scriptsize
\caption{Hyperparameters on each dataset.}
\begin{tabular}{cccc}
\toprule
Hyperparameter & Charades & ActivityNet & Description  \\  
\midrule
Learning rate   & $4 \times 10^{-4}$ &   $4 \times 10^{-4}$  & --- \\
Batch size           & 128 &   64     &   ---   \\
$n_v$           &   200     & 200     &   Number of sampled clips   \\
$n_q$           &   20      & 20      &   Number of words   \\
$d_v$           & 1,024 &   512   &   Visual feature dimension   \\
$d_w$           &   300     & 300     &   Word feature dimension  \\
$N_q$           & 1,111  &   8,000 &   Vocabulary size  \\
$d_h$           &    256    &  256    &   Hidden feature dimension  \\
%$k$            & 1  &    7       &   Number of proposals \\
$\beta_1,\beta_3$   &    0.1     &  0.1    &   Hyperparameters of $\mathcal{L}_\mathrm{IVC}$    \\ 
$\beta_2,\beta_4$   &    0.15    &  0.15   &   Hyperparameters of $\mathcal{L}_\mathrm{IVC}$     \\  
$\alpha$     &  5.5  &  5.0 &  Generating Gaussian masks \\ 
\bottomrule
\end{tabular}
%\begin{tablenotes}    
%\footnotesize                        
%\item[1] Please refer to the corresponding part for the meaning of each hyperparameter.       
%\end{tablenotes}    
\label{tab2}        
\end{table}

\section{Experiments}
In this section, we mainly evaluate the efficiency of the proposed MCMT, verify the impact of the different components, and finally perform the quantitative analysis.
\subsection{Datasets and Experimental Setting}
\subsubsection{Datasets.}
Following prior works\cite{zheng2022weakly,lin2020weakly,yang2021local,huang2021cross,chen2020look,song2020weakly}, the proposed MCMT method is evaluated on two publicly available datasets: Charades-STA (Charades) \cite{gao2017tall} and ActivityNet-Captions (ActivityNet) \cite{krishna2017dense}. Charades includes 6.7k videos primarily showcasing everyday indoor activities. It stands as the preeminent dataset for the VMR task and was initially extended and named by \cite{gao2017tall}. On the other hand, ActivityNet centers on more intricate human activities within daily life, constituting a substantial dataset for the VMR task with 20k videos. These datasets adhere to the standard partitioning established in prior research \cite{zhang2020regularized,lin2020weakly,zheng2022weakly}.
Detailed statistics of each dataset are shown in Table \ref{tab1}.

\subsubsection{Metrics.} 
To ensure impartial and rigorous comparisons, we employ the evaluation metrics ``Rank@1, $\mathrm{IoU}=m$'' and ``mIoU'' in line with previous studies \cite{chen2020look,huang2021cross,chen2021towards,zheng2022weakly}. Specifically, ``Rank@1, $\mathrm{IoU}=m$'' is defined as the percentage of natural language queries where the top-1 retrieved video moment achieves an Intersection over Union (IoU) greater than $m$. On the other hand, ``mIoU''  represents the average IoU score of the top-1 retrieved video moments across all testing queries.

\subsubsection{Implementation Details.}
Following CNM  \cite{zheng2022weakly}, we use Contrastive Language-Image Pre-training (CLIP) \cite{radford2021learning} to pre-extract visual clip features for each video in ActivityNet and Inflated 3D ConvNet (I3D) \cite{carreira2017quo} in Charades. As for the query terms, we utilize GloVe embeddings \cite{pennington2014glove} to serve as the feature extractor for the textual data. Hyperparameters are provided in Table \ref{tab2}. The transformer in the mask generator and mask conditioned reconstructor is composed of 4 attention heads and 3 layers. Note that for Charades, we limit the maximum width of the prediction to 0.45 (multiplied by Eq. \ref{eq4}) due to the shorter ground-truth length. We use Adam \cite{kingma2014adam} as the optimizer, and all experiments are conducted on dual NVIDIA GeForce RTX 2080Ti GPUs using PyTorch 1.12.1 with CUDA 11.7 and cuDNN 8.4.1.

\subsubsection{Comparison Methods}
The performance of the newly proposed MCMT method is evaluated across two benchmark datasets. Furthermore, a comparative analysis is conducted, pitting it against recently introduced methods, which encompass:

\begin{itemize}
\item{Multi-Instance Learning Methods} (MI): TGA \cite{mithun2019weakly}, WSLLN \cite{gao2019wslln}, VLANet \cite{ma2020vlanet}, WSTG \cite{chen2020look}, WSTAN \cite{wang2021weakly}, LoGAN \cite{tan2021logan}, CRM \cite{huang2021cross}, LCNet \cite{yang2021local},  

\item{Reconstruction-based Methods} (RB): WS-DEC \cite{duan2018weakly}, SCN \cite{lin2020weakly}, MARN \cite{song2020weakly}, EC-SL \cite{chen2021towards}, ACN \cite{wu2023atomic}, MARN(SA) \cite{song2023marn}, CNM,

\item{Other Weakly-supervised Method} (OW): RTBPN \cite{zhang2020regularized}.

\end{itemize}
The results on each dataset are presented in Tables \ref{tab3}-\ref{tab4}.

\begin{table*}[]
\caption{Rank@1 (\%) on the ActivityNet dataset. Con: vector concatenation, Att: additive attention mechanism. $k$: number of proposals. Values highlighted in \textbf{bold}, \underline{underlined} and \textit{italic} indicate the top-3 methods, respectively.}
%\small
% \footnotesize
\scriptsize
% \tiny
% \large
\centering
\begin{tabular}{ccccccccc}
\toprule
Type & \multicolumn{1}{c}{Model} & $\mathrm{IoU}=0.1$ & $\mathrm{IoU}=0.3$ & $\mathrm{IoU}=0.5$ & mIoU\\ \midrule
%\multirow{2}{*}{Method} & \multicolumn{3}{c|}{{Rank@1,IoU=m}} & \multicolumn{3}{c|}{Rank@5,IoU=m} \\ \cline{2-7} & 0.1 & 0.3 & 0.5 & 0.1   & 0.3 & 0.5 \\ \hline
\multirow{5}{*}{MI} & WSLLN \cite{gao2019wslln} & 75.40 & 42.80 & 22.70 & 32.20  \\
 & WSTG  \cite{chen2020look} & 74.20 & 44.30 & 23.60 & 32.20 \\
 & WSTAN \cite{wang2021weakly} & 79.78 & 52.45 & 30.01 & --- \\
 & CRM \cite{huang2021cross} & \underline{81.61} & 55.26 & 32.19 & ---  \\
 & LCNet \cite{yang2021local} & 78.58 & 48.49 & 26.33 & 34.29 \\ \midrule
 \multirow{7}{*}{RB} & WS-DEC \cite{duan2018weakly} & 62.71 & 41.98 & 23.34 & 28.23 \\
 & SCN  \cite{lin2020weakly} & 71.48 & 47.23 & 29.22 & --- \\
 & MARN  \cite{song2020weakly} & --- & 47.01 & 29.95 & --- \\
 & EC-SL \cite{chen2021towards} & 68.48 & 44.29 & 24.16 & --- \\
  & ACN \cite{wu2023atomic} & 78.78 & 57.66 & 34.18 & ---\\
 & MARN(SA) \cite{song2023marn} & --- & 48.52 & 31.37 & --- \\
 & CNM \cite{zheng2022weakly} & 78.13 & \textit{55.68} & \textit{33.33} & \textit{37.14} \\
 \midrule
\multirow{1}{*}{OW} & RTBPN \cite{zhang2020regularized} & 73.73 & 49.77 & 29.63 & --- \\ \midrule
\multirow{2}{*}{RB} & Proposed (Con, $k=12$) & \textit{81.45} & \underline{57.54} & \underline{33.56} & \underline{37.99} \\ 
& Proposed (Att, $k=7$) & \textbf{82.07} & \textbf{58.82} & \textbf{33.97} & \textbf{38.49} \\ \bottomrule
\end{tabular}
\label{tab3}
\end{table*}

\begin{table*}[]
\caption{Rank@1 (\%) on the Charades dataset.}
\scriptsize
\centering
\begin{tabular}{ccccccccc}
\toprule
%\multirow{2}{*}{Method} & \multicolumn{3}{c|}{{Rank@1,IoU=m}} & \multicolumn{3}{c|}{Rank@5,IoU=m} \\ \cline{2-7} & 0.3 & 0.5 & 0.7 & 0.3   & 0.5 & 0.7 \\ \hline
Type & Model & $\mathrm{IoU}=0.3$ & $\mathrm{IoU}=0.5$ & \multicolumn{1}{r}{$\mathrm{IoU}=0.7$} & mIoU\\ \midrule
\multirow{7}{*}{MI} & TGA \cite{mithun2019weakly} & 32.14 & 19.94 & \multicolumn{1}{r}{8.84} & --- \\
& VLANet \cite{ma2020vlanet} & 45.24 & 31.83 & \multicolumn{1}{r}{14.17} & --- \\ 
& WSTG  \cite{chen2020look} & 39.80 & 27.30 & \multicolumn{1}{r}{12.90} & 27.30\\
& WSTAN  \cite{wang2021weakly} & 43.39 & 29.35 & \multicolumn{1}{r}{12.28} & ---\\
& LoGAN \cite{tan2021logan} & 48.04 & 31.74 & \multicolumn{1}{r}{13.71} & ---\\
& CRM  \cite{huang2021cross} & 53.66 & 34.76 & \multicolumn{1}{r}{\textit{16.37}} & --- \\
& LCNet \cite{yang2021local} & 59.60 & \textbf{39.19} & \multicolumn{1}{r}{\textbf{18.87}} & \textbf{38.94} \\ \midrule
\multirow{5}{*}{RB} & SCN  \cite{lin2020weakly} & 42.96 & 23.58 & \multicolumn{1}{r}{9.97} & ---\\
& MARN  \cite{song2020weakly} & 48.55 & 31.94 & \multicolumn{1}{r}{14.81} & ---\\
& ACN \cite{wu2023atomic} & \textbf{62.95} & 37.02 & \multicolumn{1}{r}{15.26} & --- \\
& MARN(SA) \cite{song2023marn} & 48.05 & 33.87 & \multicolumn{1}{r}{15.54} & --- \\
& CNM   \cite{zheng2022weakly}  & 60.04 & 35.15 & \multicolumn{1}{r}{14.95} & 38.11 \\ 
\midrule
\multirow{1}{*}{OW} & RTBPN \cite{zhang2020regularized} & 60.04 & 32.36 & \multicolumn{1}{r}{13.24} & ---\\ \midrule
\multirow{3}{*}{RB} & Proposed (Con, $k=1$) & \underline{61.21} & 35.24 & \multicolumn{1}{r}{15.17} & \underline{38.59}\\ 
& Proposed (Att, $k=1$) & \textit{60.35} & \textit{36.23} & \multicolumn{1}{r}{15.29} & \textit{38.47} \\ 
& Proposed (Con, $k=10$ ) & 55.07 & \underline{37.65} & \multicolumn{1}{r}{\underline{18.46}} & 36.74 \\ \bottomrule
\end{tabular}
\label{tab4}
\end{table*}

\begin{table*}[h]
\caption{Rank@1 (\%) of different variants on the ActivityNet dataset. MT: Multi-task Training, MC: Multi-proposal Collaboration, Con: vector concatenation, Att: additive attention mechanism, $k$: number of proposals, $*$: the vote-based strategy during inference. Values highlighted in \textbf{bold} indicate the best variant.}
\centering
%\small
\footnotesize
%\begin{threeparttable}
\begin{tabular}{c|c|ccccc}
\toprule
\multirow{2}{*}{Row\#} & \multirow{2}{*}{Variant} & \multicolumn{5}{c}{ActivityNet} \\ 
&   & $k$ & $\mathrm{IoU}=0.1$ & $\mathrm{IoU}=0.3$ & $\mathrm{IoU}=0.5$ & mIoU \\ \midrule
    1 & \multicolumn{1}{l|}{$Baseline$} &\multicolumn{1}{r}{1} & 78.13 & 55.68 & 33.33 & 37.14 \\
    2 & \multicolumn{1}{l|}{$Baseline+MT$}  &\multicolumn{1}{r}{1} & 80.38 & 57.51 & 33.87 & 37.84  \\ \midrule
3 & \multicolumn{1}{l|}{$Baseline+MT+MC^* (Con)$}   & \multicolumn{1}{r}{12} & 81.45 & 57.54 & 33.56 & 37.99 \\ 
4 & \multicolumn{1}{l|}{$Baseline+MT+MC^* (Att)$}   & \multicolumn{1}{r}{7} & 82.07 & 58.82 & \textbf{33.97} & \textbf{38.49} \\
\midrule
5 & \multicolumn{1}{l|}{$Baseline+MT+MC (Con)$}  & \multicolumn{1}{r}{12} & 82.70 & 57.52 & 30.06 & 36.57 \\
6 & \multicolumn{1}{l|}{$Baseline+MT+MC (Att)$}  & \multicolumn{1}{r}{7} & \textbf{83.32} & \textbf{59.52} & 28.69 & 37.18 \\ \bottomrule
\end{tabular}
%\end{threeparttable}
\label{tab5}
\end{table*}

\begin{table*}[h]
\caption{Rank@1 (\%) of different variants on the Charades dataset.}
\centering
%\small
\footnotesize
%\begin{threeparttable}
\begin{tabular}{c|c|ccccc}
\toprule
\multirow{2}{*}{Row\#} & \multirow{2}{*}{Variant} & \multicolumn{5}{c}{Charades} \\ 
&  & \multicolumn{1}{r}{$k$}  & $\mathrm{IoU}=0.3$ & $\mathrm{IoU}=0.5$ & $\mathrm{IoU}=0.7$ &mIoU \\ \midrule
    1 & \multicolumn{1}{l|}{$Baseline$} & \multicolumn{1}{r}{1} & 60.04 & 35.15 & 14.95 & 38.11  \\
    2 & \multicolumn{1}{l|}{$Baseline+MT$} & \multicolumn{1}{r}{1} & \textbf{61.21} & 35.24 & 15.17 & \textbf{38.59} \\ \midrule
3 & \multicolumn{1}{l|}{$Baseline+MT+MC^* (Con)$}  & \multicolumn{1}{r}{10} & 55.07 & \textbf{37.65} & \textbf{18.46} & 36.74 \\ 
4 & \multicolumn{1}{l|}{$Baseline+MT+MC^* (Att)$}  & \multicolumn{1}{r}{9} & 55.64 & 37.11 & 17.00 & 36.96  \\
\midrule
5 & \multicolumn{1}{l|}{$Baseline+MT+MC (Con)$} & \multicolumn{1}{r}{10} & 54.72 & \textbf{37.65} & 18.08 & 36.68  \\
6 & \multicolumn{1}{l|}{$Baseline+MT+MC (Att)$}  & \multicolumn{1}{r}{9} & 54.84 & 36.89 & 16.72 & 36.64  \\ \bottomrule
\end{tabular}
%\end{threeparttable}
\label{tab6}
\end{table*}

\subsection{Experimental Analysis}
\subsubsection{Overall Performance Comparison}

We first compare the proposed MCMT on the ActivityNet dataset with the recent methods in Table \ref{tab3}. MCMT achieves the best results across all metrics. Specifically, it outperforms the recent multi-instance learning methods consistently. 
Notably, CRM introduces additional paragraph annotation information during training, while MCMT achieves superior performance without such information.
For reconstruction-based and other weakly-supervised methods, MCMT also shows better results, particularly compared to the baseline reconstruction-based method CNM, , which is of the same type as MCMT. MCMT surpasses CNM with improvements of 3.94\%, 3.14\%, 0.64\%, and 1.35\% across various metrics. These results suggest that the multi-proposal collaboration and multi-task training mechanisms are highly effective in scenarios where complex activities with diverse temporal structures need to be captured.

To further elaborate on these improvements, MCMT's advantage is particularly noticeable when dealing with long-duration videos containing multiple activities, as is common in the ActivityNet dataset. The multi-task training mechanism allows MCMT to maintain better alignment between queries and video moments through forward and inverse query reconstruction tasks, effectively constraining the model from multiple perspectives. This dual-constraint strategy strengthens the model's robustness and ability to generalize across various video lengths and complexities, resulting in improved temporal localization accuracy.

Additionally, MCMT’s multiple proposals improve performance, especially at challenging IoU thresholds like IoU = 0.3 and IoU = 0.5. This indicates that the model can handle finer temporal localization with higher precision, particularly in videos with complex temporal dynamics. Compared to single-proposal methods, MCMT reduces the likelihood of redundant information from neighboring video segments influencing the retrieval results, thereby enhancing its capability in both coarse and fine retrieval tasks.

For the Charades dataset, MCMT ranks among the top-3 across all metrics (Table \ref{tab4}). While it demonstrates minor deviations from other methods in some metrics, these differences are not substantial. Specifically, MCMT achieves the second-best result on the $\mathrm{IoU}=0.3$ metric, with a 1.17\% improvement over both the baseline reconstruction-based method CNM and the other weakly-supervised method RTBPN \cite{zhang2020regularized}. This shows that MCMT is particularly strong in coarse retrieval tasks. However, as we increase the number of proposals, we observe slight improvements at higher IoU thresholds ($\mathrm{IoU}=0.5$, $\mathrm{IoU}=0.7$), though still falling short of the LCNet \cite{yang2021local} method. The reason for this may be attributed to LCNet’s multi-instance learning strategy, which focuses on fine-grained correspondences between video and text via self-supervised learning. In contrast, MCMT is a reconstruction-based method. These two methods guide model training from different perspectives, leading to the observed performance differences. Nonetheless, MCMT still outperforms other reconstruction-based methods on nearly all metrics, emphasizing the effectiveness of the proposed method.

Upon closer examination, we observe that MCMT encounters challenges in distinguishing fine-grained actions within highly similar indoor environments, such as those in the Charades dataset. When activities share overlapping backgrounds and objects, the multi-proposal mechanism can struggle to differentiate subtle temporal transitions between actions, leading to slightly lower scores at higher IoU thresholds (e.g., IoU = 0.5). This issue is particularly prominent in the Charades dataset, where background similarity poses a significant challenge for proposal-based models. Despite this limitation, MCMT remains competitive across a wide range of IoU thresholds, which demonstrates the model's robustness in handling coarse-level moment retrieval, especially in scenes with diverse activity types and temporal structures.

Moreover, considering the different data sources in the two datasets, the ActivityNet dataset contains more diverse and flexible human activities, while Charades focuses on daily indoor activities with similar backgrounds, making the latter dataset more challenging. These differences cause the model to perform differently on each dataset. In ActivityNet, the diversity in temporal structures and activity types provides a broader range for MCMT to fully utilize its multi-proposal mechanism. In contrast, the similarity in Charades limits the model's ability to exploit this mechanism to its fullest extent. Nevertheless, the strong performance in both datasets, especially in coarse-grained retrieval tasks, highlights the robustness of the MCMT model in handling various video types.

\subsubsection{Discussion}
We systematically analyze the limitations of our study and their impact on the interpretation of the results. Specifically, experimental results show that at an IoU threshold of 0.5, the Rank@1 of MCMT on the Charades dataset is 37.65\%, slightly below LCNet’s~\cite{yang2021local} 39.19\% in Table~\ref{tab4}. This performance difference can primarily be attributed to the inherent characteristics of the dataset and the design of the method: the Charades dataset features a large number of highly similar indoor activities, making fine-grained action distinctions particularly challenging. For example, actions like ``picking up a book" and ``putting down a book" differ only slightly in temporal sequence, with nearly identical backgrounds. Additionally, while the model utilizes multi-proposal collaboration to generate candidate proposals, the impact of increasing the proposal quantity on performance at higher IoU thresholds (e.g., IoU=0.7) is limited in the Charades dataset. This highlights how redundant information and background similarity constrain the precision of the proposal module. Nevertheless, the method still outperforms other reconstruction-based methods (e.g., CNM~\cite{zheng2022weakly}) across several metrics, demonstrating the robustness and generalizability of MCMT in the weakly-supervised video moment retrieval task. By integrating these findings, we have provided a detailed analysis of the performance gaps and cautiously interpreted the applicability of the results.

\begin{figure*}
\centering
\subfigure[ActivityNet (Att: Additive attention mechanism)]{
\begin{minipage}[b]{0.47\textwidth}
\includegraphics[width=1\textwidth]{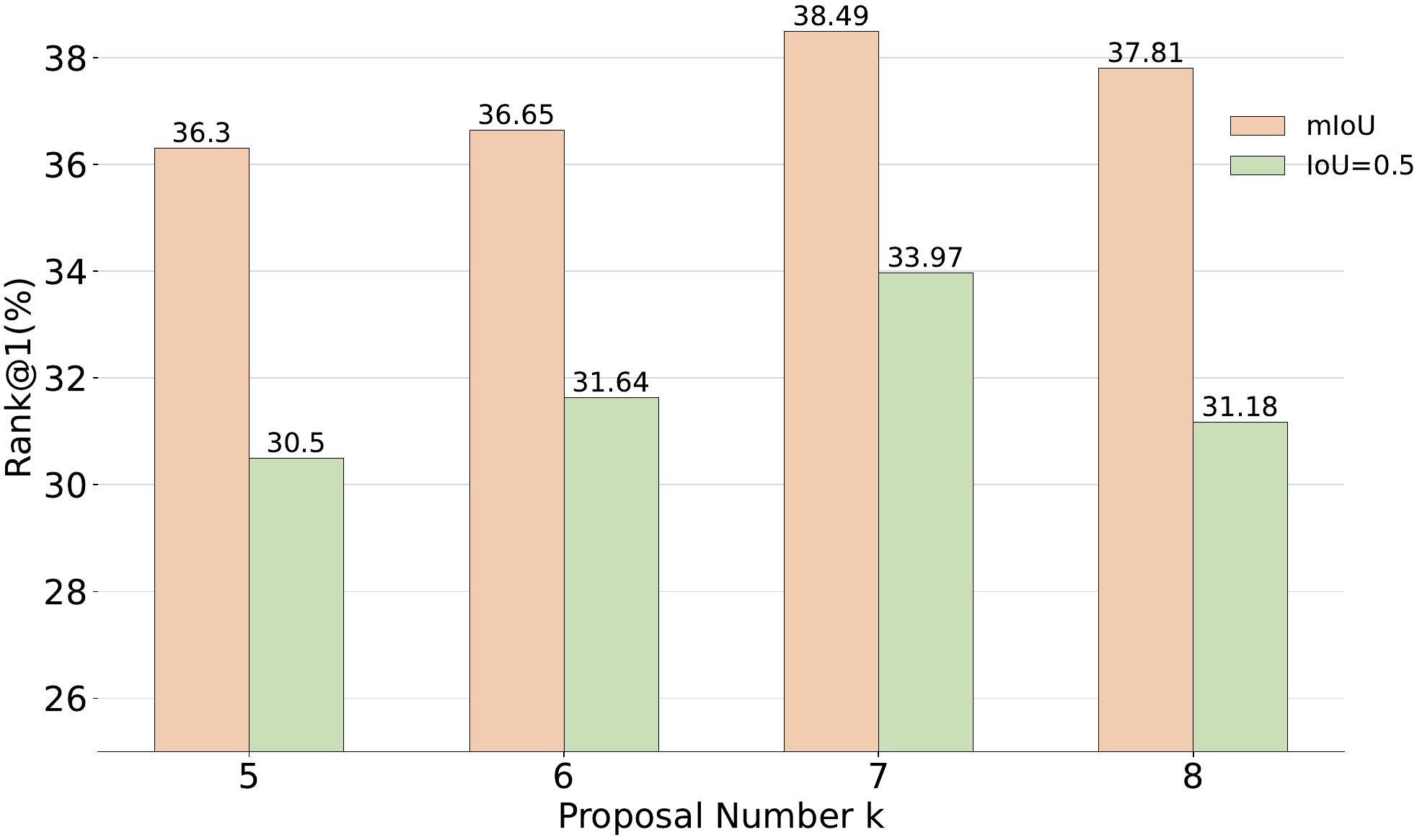} 
\end{minipage}
}
\subfigure[Charades (Con: Vector concatenation)]{
\begin{minipage}[b]{0.47\textwidth}
\includegraphics[width=1.0\textwidth]{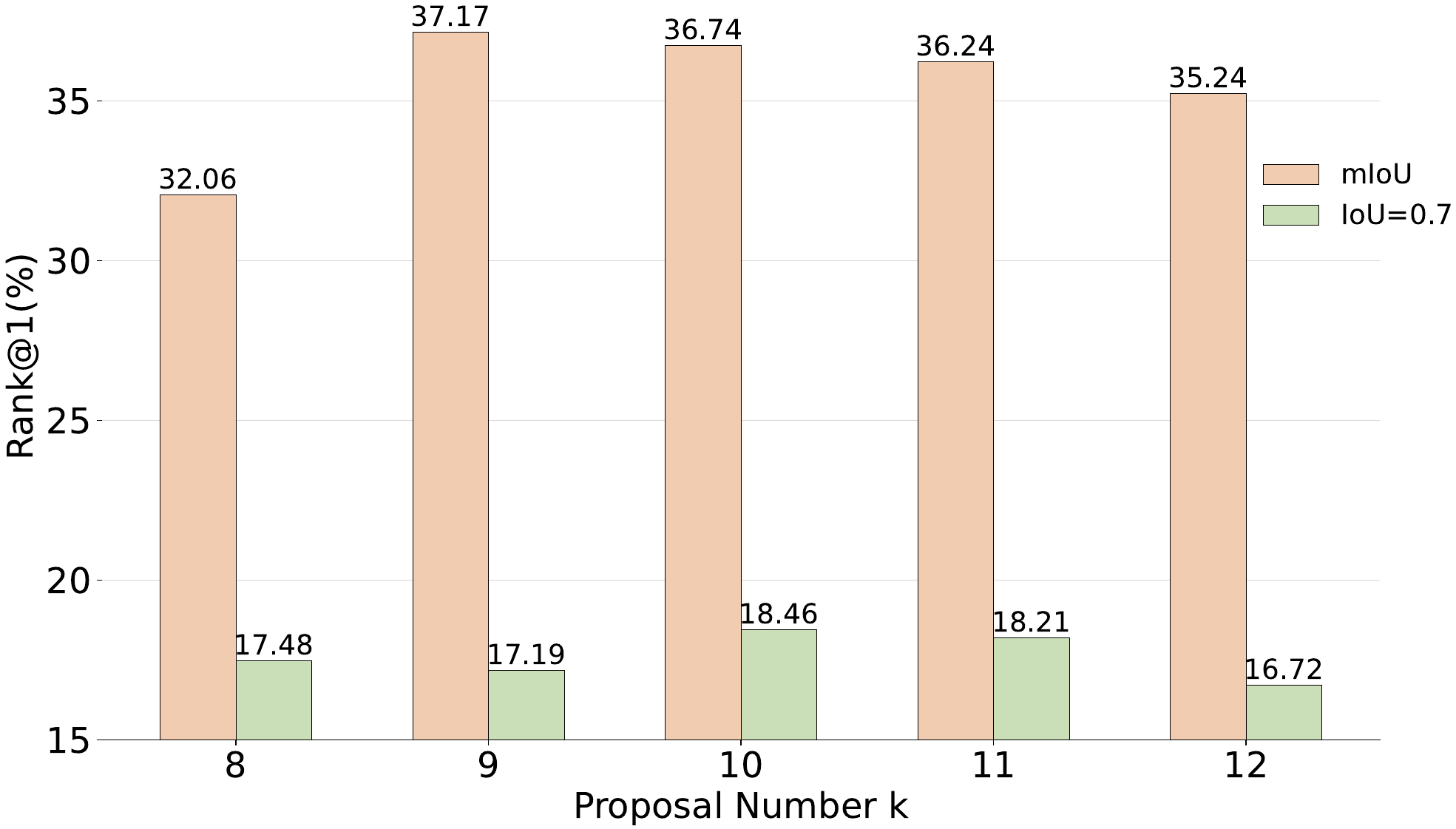} 
\end{minipage}
}
\centering
\caption{Effect of the proposal number $k$ on each dataset. }
\label{fig3}
\end{figure*}

\subsection{Ablation Studies}
Furthermore, we undertake ablation studies to explore the influence of various factors on the performance of the proposed MCMT.
\subsubsection{Efficiency of the Proposed Components}
Tables \ref{tab5}-\ref{tab6} show the evaluation results of the different variants on the two datasets. Since CNM \cite{zheng2022weakly} is trained and generates a single proposal under the guidance of the forward masked query reconstruction task, we treat it as the Baseline. As can be seen from Rows 1--2 in Tables \ref{tab5}-\ref{tab6}, after introducing the Multi-task Training ($MT$) mechanism, $Baseline+MT$ performs significantly better than Baseline on both datasets, which validates the effectiveness of $MT$ by imposing substantial constraints and guidance on the model. However, the performance of the variants on the two datasets is inconsistent when the Multi-proposal Collaboration ($MC$) mechanism is introduced. Observing Rows 2--4 of Table \ref{tab5} on the ActivityNet dataset, the performance of $Baseline+MT+MC^* (Con)$ and $Baseline+MT+MC^* (Att)$ is better than that of $Baseline+MT$. In particular, the variant with $MC (Att)$ shows the best performance, which validates the effectiveness of the $MC$ mechanism on this dataset. Meanwhile, observing Rows 2--4 of Table \ref{tab6} on the Charades dataset, when we introduce the $MC$ mechanism, the evaluation results of the corresponding variants decrease on $\mathrm{IoU}=0.3$ and mIoU, but improve greatly on $\mathrm{IoU}=0.5$ and $\mathrm{IoU}=0.7$, which indicates that the $MC$ mechanism enables the model to achieve finer localization of video moments. In conclusion, these experiments verify the effectiveness of the $MT$ and $MC$ mechanisms.
\subsubsection{Vector Concatenation v.s. Additive Attention Mechanism}
Vector concatenation and additive attention mechanism are two ways used for multi-modal information fusion in our method. We compare their performance on the two datasets, as shown in Tables \ref{tab3}-\ref{tab4} (Proposed (Con) v.s. Proposed (Att)). We observe that additive attention mechanism (Att) performs better than vector concatenation (Con) on the ActivityNet dataset while comparable on the Charades dataset. We recommend adopting additive attention operation since it achieves better performance with fewer parameters.

\subsubsection{Attention Score Strategy v.s. Vote-based Strategy}
Two strategies during inference are compared. Observing Rows 3--6 of Tables \ref{tab5}-\ref{tab6}, we notice that the variants employing the two strategies show similar performance  in the coarser evaluation metrics, but the variants with the vote-based strategy significantly outperform the variants with the attention score strategy on the finer evaluation metrics. In particular, the variant based on the voting strategy improves at least 3.5\% on $\mathrm{IoU}=0.5$ on the ActivityNet dataset. In summary, we recommend the use of the vote-based strategy.

\subsubsection{Number of Generated Proposals}
We further explore the effect of the number of generated proposals on model retrieval performance. We vary the number of generated proposals and report the experimental results in Figure \ref{fig3}. We select mIoU as the model's global performance evaluation metric on both datasets, and choose $\mathrm{IoU}=0.5$ and $\mathrm{IoU}=0.7$ as more stringent evaluation metrics on the ActivityNet and Charades datasets, respectively.  We notice that on the ActivityNet dataset, with the rise of $k$, the model's performance first increases and then decreases, and it is optimal when $k=7$.  Meanwhile, on the Charades dataset, 
when we continue to increase the number of proposals, the overall performance of the model (mIoU) also shows a trend of increasing first and then decreasing.
As for the more stringent retrieval performance ($\mathrm{IoU}=0.7$), when $k=10$, the performance of the model is optimal.
Moreover, the impact trend shows that the number of generated proposals has different impacts on different datasets due to the different specific data distributions.

\begin{table}[t]
    \centering
    \caption{Impact of Hard Negative Masks vs. Easy Negative Masks on Charades dataset.}
    \begin{tabular}{c|cc|cccc}
        \toprule
        Row\# & Hard & Easy & IoU=0.3 & IoU=0.5 & IoU=0.7 & mIoU \\
        \midrule
        1 & \checkmark & \checkmark & \textbf{55.07} & \textbf{37.65} & \textbf{18.46} & \textbf{36.74} \\
        2& $\times$ & \checkmark & 52.28 & 32.96 & 15.58 & 33.90 \\        
        3 & \checkmark & $\times$ & 54.65 & 36.26 & 16.69 & 35.75 \\
        
        4 & $\times$ & $\times$ & 40.69 & 29.58 & 14.19 & 27.90 \\
        \bottomrule
    \end{tabular}
    \label{hvse}
\end{table}

\subsubsection{Impact of Hard Negative Masks vs. Easy Negative Masks}
From the results in Table \ref{hvse}, it is clear that the model with both hard and easy negative samples (Row 1) achieves the best performance in almost all metrics. This indicates that the combination of hard and easy negative samples enables the model to learn more diverse negative examples, which in turn improves its ability to distinguish between positive and negative samples. Hard negatives provide more challenging examples, while easy negatives help the model quickly learn basic discriminative abilities. The combination of both types maximizes the model's learning capacity, especially in complex scenarios requiring higher generalization.

In contrast, models using only easy negatives (Row 2) or only hard negatives (Row 3) show inferior performance, although hard negatives slightly outperform easy negatives at these metrics, but the overall performance still lags behind the combination of both types. Notably, at IoU=0.7, the model with hard negatives slightly outperforms the one with easy negatives (16.69 vs. 15.58). Furthermore, the model without any negative samples (Row 4) performs the worst, with an mIoU of only 27.90, highlighting the critical role of negative samples in improving the model’s discriminative power.

\begin{table}[t]
    \centering
    \caption{Impact of Forward Queries vs. Inverse Queries on Charades dataset.}
    \begin{tabular}{c|l|cccc}
        \toprule
        Row\# & Variant & IoU=0.3 & IoU=0.5 & IoU=0.7 & mIoU \\
        \midrule
        1 & MCMT w/ Forward Queries & 52.34 & 36.67 & 17.76 & 35.01 \\
        
        2 & MCMT w/ Inverse Queries & \textbf{55.13} & 37.21 & 17.70 & 36.34 \\
        3 & MCMT w/ All Queries & 55.07 & \textbf{37.65} & \textbf{18.46} & \textbf{36.74} \\
        \bottomrule
    \end{tabular}
    \label{fvsi}
\end{table}

\subsubsection{Impact of Forward Queries vs. Inverse Queries}
From the results in Table \ref{fvsi}, the model with both forward and inverse queries combined (Row 3) achieves the highest mIoU of 36.74 and the best performance at IoU=0.5 and IoU=0.7, with scores of 37.65 and 18.46, respectively. This demonstrates that combining both forward and inverse queries provides the model with richer training signals, enhancing its contrastive learning from multiple perspectives and improving overall performance. This multi-view training signal is particularly important for the model’s performance in complex scenarios, especially at higher IoU thresholds.

In comparison, the model using only forward queries (Row 1) shows a slight decline in performance, indicating that forward queries alone may not provide enough constraint for the model in some cases. The model using only inverse queries (Row 2) achieves the best IoU=0.3 score (55.13) but lags behind in other metrics, particularly at IoU=0.7 (17.70). Overall, the combination of both forward and inverse queries offers a more comprehensive supervision signal, resulting in improved overall performance.

\subsubsection{Impact of Different Loss Functions}
Regarding the exploration of different loss functions, we have already addressed this in previous ablation studies. For example, in Table~\ref{fvsi}, Row 1 shows the retrieval performance of the model using only 
$\mathcal{L}_\mathrm{CE\text{-}f}^p$, $\mathcal{L}_\mathrm{CE\text{-}f}^h$, and $\mathcal{L}_\mathrm{IVC\text{-}f}$, focusing specifically on the contribution of these loss functions. Similarly, Table~\ref{hvse} presents the results related to Hard and Easy Negatives. Row 2 in Table~\ref{hvse} illustrates the model's performance when the hard negative reconstruction loss and hard intra-video loss are excluded, leaving only the easy intra-video loss. These analyses already cover the impact of different loss functions in various scenarios. We encourage you to review these sections for further details. Overall, the proposed method has demonstrated its effectiveness through these ablation studies.

\begin{figure}
\centering
\subfigure[Charades]{
\begin{minipage}[b]{0.75\textwidth}
\includegraphics[width=1.0\textwidth]{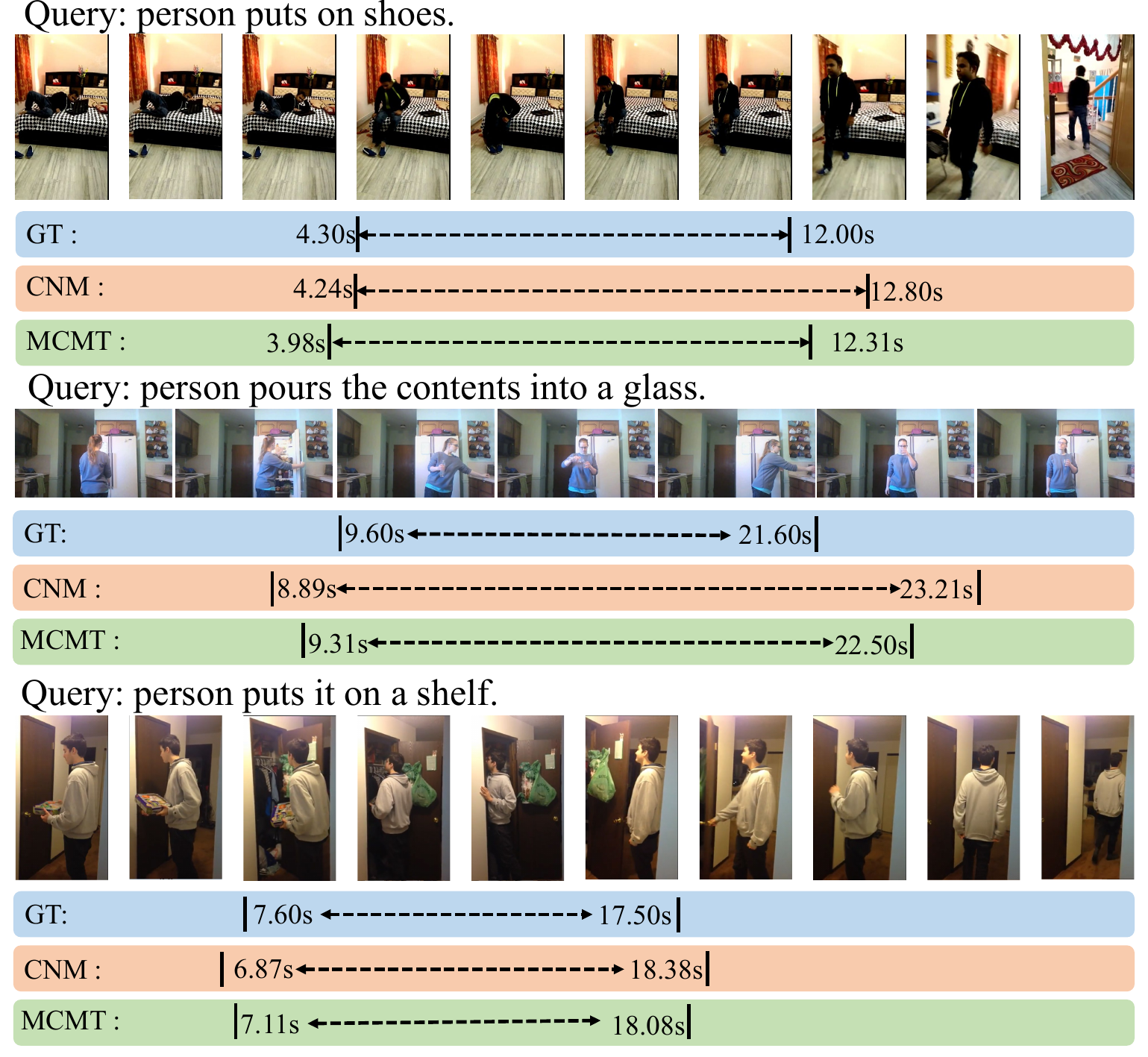} 
\end{minipage}
}
\subfigure[ActivityNet]{
\begin{minipage}[b]{0.75\textwidth}
\includegraphics[width=1.0\textwidth]{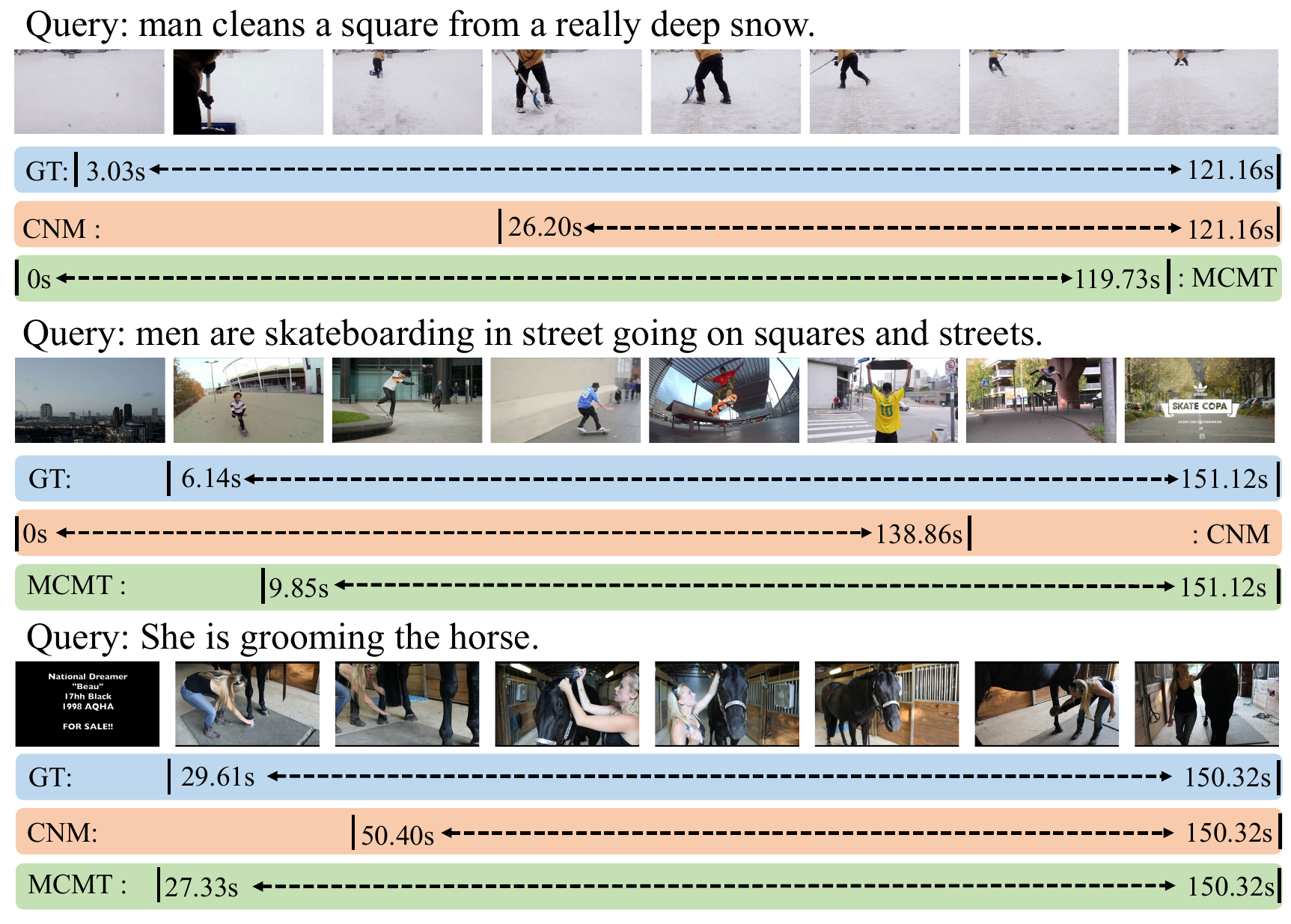} 
\end{minipage}
}
\centering
\caption{Qualitative examples of each dataset.}
\label{fig4}
\end{figure}

\subsection{Qualitative Results}
In order to qualitatively confirm the effectiveness of the proposed MCMT method, we present some typical examples illustrating moment retrieval guided by natural language.  Figure \ref{fig4} shows the retrieval results achieved by the MCMT method as well as the Baseline CNM \cite{zheng2022weakly} on the Charades and ActivityNet datasets, respectively. Through intuitive comparison, MCMT is closer to the ground truth than CNM. Unlike CNM which directly generates only a Gaussian mask as the positive mask and introduces a single task to guide model training, MCMT aggregates the information of multiple predicted Gaussian masks to generate a robust positive mask and mine negative masks based on this high-quality positive mask. In addition to the forward reconstruction masked query task, MCMT introduces the inverse reconstruction masked query task to impose more substantial constraints on the model to guide model training. All these initiatives make our models more efficient and stable.

\subsection{Lessons Learned}
We further analyze and summarize the findings revealed by the experimental results. First, comparing Table~\ref{tab3} and Table~\ref{tab4}, we find that the proposed MCMT achieves optimal values on all metrics in the ActivityNet dataset, while it achieves competitive results in the Charades dataset. Considering the distinct data sources between the two datasets, the videos within the ActivityNet dataset record flexible human activities, exhibiting diversity and openness. On the other hand, the videos in the Charades dataset focus on recording indoor daily activities, with similar backgrounds. This variance in data distributions contributes to different performances. Specifically, this is the main reason that the proposed method is less effective on Charades compared to ActivityNet. Second, comparing the closest method CNM \cite{zheng2022weakly} as the Baseline, observing Rows 1--2 of Tables \ref{tab5}-\ref{tab6}, the method variant outperforms CNM across the board when the $MT$ mechanism is introduced. Comparing Rows 1 and 3, the performance of the variant improves on the ActivityNet dataset when the $MC$ mechanism is introduced, and interestingly on the Charades dataset, $\mathrm{IoU}=0.3$ and mIoU decrease while $\mathrm{IoU}=0.5$ and $\mathrm{IoU}=0.7$ improve, which indicates that the $MC$ mechanism makes the model focus on more accurate moment retrieval. Taken together, experiments and ablation studies as well as quantitative results show the effectiveness of the proposed MCMT.

\section{CONCLUSION}
This paper presents a novel approach, called \textbf{M}ulti-proposal \textbf{C}ollaboration and \textbf{M}ulti-task \textbf{T}raining (MCMT), for the weakly-supervised Video Moment Retrieval (VMR) task. MCMT generates multiple Gaussian masks and uses a mask aggregation module to create a robust final positive sample mask, effectively mining high-quality positive, easy negative, and hard negative samples. In addition, MCMT introduces a dual-task mechanism, where the forward and reverse masked query reconstruction tasks impose stronger constraints on the model, enhancing performance during training. Experimental results on two widely-used datasets, Charades and ActivityNet, demonstrate the superiority of MCMT, where it achieves significant performance improvements.

In the future, we plan to incorporate more flexible cross-modal attention mechanisms to improve the model’s ability to capture fine-grained actions. Additionally, we aim to optimize the proposal generation module by integrating clustering strategies to enhance proposal quality and reduce redundancy. To further improve the model's generalization, we will explore data augmentation techniques to create more diverse training samples. These directions are expected to further enhance the model's performance and applicability, offering more possibilities for the weakly-supervised video moment retrieval task.

\section*{Acknowledgments}
This work was supported in part by the China Scholarship Council under No.202206130025, Postgraduate Scientific Research Innovation Project of Hunan Province under Grant QL20220096, National Natural Science Foundation of China under Grant 62072169, 62172156, Natural Science Foundation of Hunan Province under Grant 2021JJ30152.

\section*{Declarations}
\bmhead{Authors contribution statement} Mr.Zhang designed the framework, performed the experiments and wrote the manuscript. Ms.Yang assisted with the implementation of the research, verification of the experimental results and revised the manuscript. Mr.Jiang, Mr.Komamizu and Mr.IDE provided
critical feedback and contributed to the final manuscript.
\bmhead{Data availability and access} The raw data elaborated during the current research are publicly available.
\bmhead{Competing Interests} The authors declare that they have no competing
interests.
\bmhead{Ethical and informed consent for data used} Not applicable.

\bibliography{sn-bibliography.bib}% common bib file


\begin{thebibliography}{41}
\ifx \bisbn   \undefined \def \bisbn  #1{ISBN #1}\fi
\ifx \binits  \undefined \def \binits#1{#1}\fi
\ifx \bauthor  \undefined \def \bauthor#1{#1}\fi
\ifx \batitle  \undefined \def \batitle#1{#1}\fi
\ifx \bjtitle  \undefined \def \bjtitle#1{#1}\fi
\ifx \bvolume  \undefined \def \bvolume#1{\textbf{#1}}\fi
\ifx \byear  \undefined \def \byear#1{#1}\fi
\ifx \bissue  \undefined \def \bissue#1{#1}\fi
\ifx \bfpage  \undefined \def \bfpage#1{#1}\fi
\ifx \blpage  \undefined \def \blpage #1{#1}\fi
\ifx \burl  \undefined \def \burl#1{\textsf{#1}}\fi
\ifx \doiurl  \undefined \def \doiurl#1{\url{https://doi.org/#1}}\fi
\ifx \betal  \undefined \def \betal{\textit{et al.}}\fi
\ifx \binstitute  \undefined \def \binstitute#1{#1}\fi
\ifx \binstitutionaled  \undefined \def \binstitutionaled#1{#1}\fi
\ifx \bctitle  \undefined \def \bctitle#1{#1}\fi
\ifx \beditor  \undefined \def \beditor#1{#1}\fi
\ifx \bpublisher  \undefined \def \bpublisher#1{#1}\fi
\ifx \bbtitle  \undefined \def \bbtitle#1{#1}\fi
\ifx \bedition  \undefined \def \bedition#1{#1}\fi
\ifx \bseriesno  \undefined \def \bseriesno#1{#1}\fi
\ifx \blocation  \undefined \def \blocation#1{#1}\fi
\ifx \bsertitle  \undefined \def \bsertitle#1{#1}\fi
\ifx \bsnm \undefined \def \bsnm#1{#1}\fi
\ifx \bsuffix \undefined \def \bsuffix#1{#1}\fi
\ifx \bparticle \undefined \def \bparticle#1{#1}\fi
\ifx \barticle \undefined \def \barticle#1{#1}\fi
\bibcommenthead
\ifx \bconfdate \undefined \def \bconfdate #1{#1}\fi
\ifx \botherref \undefined \def \botherref #1{#1}\fi
\ifx \url \undefined \def \url#1{\textsf{#1}}\fi
\ifx \bchapter \undefined \def \bchapter#1{#1}\fi
\ifx \bbook \undefined \def \bbook#1{#1}\fi
\ifx \bcomment \undefined \def \bcomment#1{#1}\fi
\ifx \oauthor \undefined \def \oauthor#1{#1}\fi
\ifx \citeauthoryear \undefined \def \citeauthoryear#1{#1}\fi
\ifx \endbibitem  \undefined \def \endbibitem {}\fi
\ifx \bconflocation  \undefined \def \bconflocation#1{#1}\fi
\ifx \arxivurl  \undefined \def \arxivurl#1{\textsf{#1}}\fi
\csname PreBibitemsHook\endcsname

%%% 1
\bibitem[\protect\citeauthoryear{Collins et~al.}{2000}]{collins2000system}
\begin{barticle}
\bauthor{\bsnm{Collins}, \binits{R.T.}},
\bauthor{\bsnm{Lipton}, \binits{A.J.}},
\bauthor{\bsnm{Kanade}, \binits{T.}},
\bauthor{\bsnm{Fujiyoshi}, \binits{H.}},
\bauthor{\bsnm{Duggins}, \binits{D.}},
\bauthor{\bsnm{Tsin}, \binits{Y.}},
\bauthor{\bsnm{Tolliver}, \binits{D.}},
\bauthor{\bsnm{Enomoto}, \binits{N.}},
\bauthor{\bsnm{Hasegawa}, \binits{O.}},
\bauthor{\bsnm{Burt}, \binits{P.}},
\bauthor{\bsnm{Wixson}, \binits{L.}}:
\batitle{A system for video surveillance and monitoring}.
\bjtitle{VSAM Final Report}
\bvolume{2000},
\bfpage{1}--\blpage{68}
(\byear{2000})
\end{barticle}
\endbibitem

%%% 2
\bibitem[\protect\citeauthoryear{He et~al.}{2024}]{he2024video}
\begin{botherref}
\oauthor{\bsnm{He}, \binits{Q.}},
\oauthor{\bsnm{Shi}, \binits{R.}},
\oauthor{\bsnm{Chen}, \binits{L.}},
\oauthor{\bsnm{Huo}, \binits{L.}}:
Video anomaly detection based on multi-scale optical flow spatio-temporal
  enhancement and normality mining.
International Journal of Machine Learning and Cybernetics,
1--16
(2024)
\end{botherref}
\endbibitem

%%% 3
\bibitem[\protect\citeauthoryear{Kemp et~al.}{2007}]{kemp2007challenges}
\begin{barticle}
\bauthor{\bsnm{Kemp}, \binits{C.C.}},
\bauthor{\bsnm{Edsinger}, \binits{A.}},
\bauthor{\bsnm{Torres-Jara}, \binits{E.}}:
\batitle{Challenges for robot manipulation in human environments [grand
  challenges of robotics]}.
\bjtitle{IEEE Robotics \& Automation Magazine}
\bvolume{14},
\bfpage{20}--\blpage{29}
(\byear{2007})
\end{barticle}
\endbibitem

%%% 4
\bibitem[\protect\citeauthoryear{Anne~Hendricks
  et~al.}{2017}]{anne2017localizing}
\begin{bchapter}
\bauthor{\bsnm{Anne~Hendricks}, \binits{L.}},
\bauthor{\bsnm{Wang}, \binits{O.}},
\bauthor{\bsnm{Shechtman}, \binits{E.}},
\bauthor{\bsnm{Sivic}, \binits{J.}},
\bauthor{\bsnm{Darrell}, \binits{T.}},
\bauthor{\bsnm{Russell}, \binits{B.}}:
\bctitle{Localizing moments in video with natural language}.
In: \bbtitle{Proceedings of the 16th IEEE International Conference on Computer
  Vision},
pp. \bfpage{5803}--\blpage{5812}
(\byear{2017})
\end{bchapter}
\endbibitem

%%% 5
\bibitem[\protect\citeauthoryear{Gao et~al.}{2017}]{gao2017tall}
\begin{bchapter}
\bauthor{\bsnm{Gao}, \binits{J.}},
\bauthor{\bsnm{Sun}, \binits{C.}},
\bauthor{\bsnm{Yang}, \binits{Z.}},
\bauthor{\bsnm{Nevatia}, \binits{R.}}:
\bctitle{Tall: Temporal activity localization via language query}.
In: \bbtitle{Proceedings of the 16th IEEE International Conference on Computer
  Vision},
pp. \bfpage{5267}--\blpage{5275}
(\byear{2017})
\end{bchapter}
\endbibitem

%%% 6
\bibitem[\protect\citeauthoryear{Zhang et~al.}{2022a}]{zhang2022dual}
\begin{bchapter}
\bauthor{\bsnm{Zhang}, \binits{B.}},
\bauthor{\bsnm{Jiang}, \binits{B.}},
\bauthor{\bsnm{Yang}, \binits{C.}},
\bauthor{\bsnm{Pang}, \binits{L.}}:
\bctitle{Dual-channel localization networks for moment retrieval with natural
  language}.
In: \bbtitle{Proceedings of the 2022 International Conference on Multimedia
  Retrieval},
pp. \bfpage{351}--\blpage{359}
(\byear{2022})
\end{bchapter}
\endbibitem

%%% 7
\bibitem[\protect\citeauthoryear{Zhang et~al.}{2022b}]{zhang2022video}
\begin{bchapter}
\bauthor{\bsnm{Zhang}, \binits{B.}},
\bauthor{\bsnm{Yang}, \binits{C.}},
\bauthor{\bsnm{Jiang}, \binits{B.}},
\bauthor{\bsnm{Zhou}, \binits{X.}}:
\bctitle{Video moment retrieval with hierarchical contrastive learning}.
In: \bbtitle{Proceedings of the 30th ACM International Conference on
  Multimedia},
pp. \bfpage{346}--\blpage{355}
(\byear{2022})
\end{bchapter}
\endbibitem

%%% 8
\bibitem[\protect\citeauthoryear{Liu et~al.}{2018}]{liu2018attentive}
\begin{bchapter}
\bauthor{\bsnm{Liu}, \binits{M.}},
\bauthor{\bsnm{Wang}, \binits{X.}},
\bauthor{\bsnm{Nie}, \binits{L.}},
\bauthor{\bsnm{He}, \binits{X.}},
\bauthor{\bsnm{Chen}, \binits{B.}},
\bauthor{\bsnm{Chua}, \binits{T.-S.}}:
\bctitle{Attentive moment retrieval in videos}.
In: \bbtitle{The 41st International ACM SIGIR Conference on Research \&
  Development in Information Retrieval},
pp. \bfpage{15}--\blpage{24}
(\byear{2018})
\end{bchapter}
\endbibitem

%%% 9
\bibitem[\protect\citeauthoryear{Wang et~al.}{2022}]{wang2022siamese}
\begin{barticle}
\bauthor{\bsnm{Wang}, \binits{Y.}},
\bauthor{\bsnm{Liu}, \binits{M.}},
\bauthor{\bsnm{Wei}, \binits{Y.}},
\bauthor{\bsnm{Cheng}, \binits{Z.}},
\bauthor{\bsnm{Wang}, \binits{Y.}},
\bauthor{\bsnm{Nie}, \binits{L.}}:
\batitle{Siamese alignment network for weakly supervised video moment
  retrieval}.
\bjtitle{IEEE Transactions on Multimedia}
\bvolume{25},
\bfpage{3921}--\blpage{3933}
(\byear{2022})
\end{barticle}
\endbibitem

%%% 10
\bibitem[\protect\citeauthoryear{Yoon et~al.}{2023}]{yoon2023scanet}
\begin{bchapter}
\bauthor{\bsnm{Yoon}, \binits{S.}},
\bauthor{\bsnm{Koo}, \binits{G.}},
\bauthor{\bsnm{Kim}, \binits{D.}},
\bauthor{\bsnm{Yoo}, \binits{C.D.}}:
\bctitle{Scanet: Scene complexity aware network for weakly-supervised video
  moment retrieval}.
In: \bbtitle{Proceedings of the IEEE/CVF International Conference on Computer
  Vision},
pp. \bfpage{13576}--\blpage{13586}
(\byear{2023})
\end{bchapter}
\endbibitem

%%% 11
\bibitem[\protect\citeauthoryear{Huang et~al.}{2023}]{huang2023weakly}
\begin{bchapter}
\bauthor{\bsnm{Huang}, \binits{Y.}},
\bauthor{\bsnm{Yang}, \binits{L.}},
\bauthor{\bsnm{Sato}, \binits{Y.}}:
\bctitle{Weakly supervised temporal sentence grounding with uncertainty-guided
  self-training}.
In: \bbtitle{Proceedings of the IEEE/CVF Conference on Computer Vision and
  Pattern Recognition},
pp. \bfpage{18908}--\blpage{18918}
(\byear{2023})
\end{bchapter}
\endbibitem

%%% 12
\bibitem[\protect\citeauthoryear{Lv et~al.}{2023}]{lv2023counterfactual}
\begin{bchapter}
\bauthor{\bsnm{Lv}, \binits{Z.}},
\bauthor{\bsnm{Su}, \binits{B.}},
\bauthor{\bsnm{Wen}, \binits{J.-R.}}:
\bctitle{Counterfactual cross-modality reasoning for weakly supervised video
  moment localization}.
In: \bbtitle{Proceedings of the 31st ACM International Conference on
  Multimedia},
pp. \bfpage{6539}--\blpage{6547}
(\byear{2023})
\end{bchapter}
\endbibitem

%%% 13
\bibitem[\protect\citeauthoryear{Mithun et~al.}{2019}]{mithun2019weakly}
\begin{bchapter}
\bauthor{\bsnm{Mithun}, \binits{N.C.}},
\bauthor{\bsnm{Paul}, \binits{S.}},
\bauthor{\bsnm{Roy-Chowdhury}, \binits{A.K.}}:
\bctitle{Weakly supervised video moment retrieval from text queries}.
In: \bbtitle{Proceedings of the 2019 IEEE/CVF Conference on Computer Vision and
  Pattern Recognition},
pp. \bfpage{11592}--\blpage{11601}
(\byear{2019})
\end{bchapter}
\endbibitem

%%% 14
\bibitem[\protect\citeauthoryear{Tan et~al.}{2021}]{tan2021logan}
\begin{bchapter}
\bauthor{\bsnm{Tan}, \binits{R.}},
\bauthor{\bsnm{Xu}, \binits{H.}},
\bauthor{\bsnm{Saenko}, \binits{K.}},
\bauthor{\bsnm{Plummer}, \binits{B.A.}}:
\bctitle{Logan: Latent graph co-attention network for weakly-supervised video
  moment retrieval}.
In: \bbtitle{Proceedings of the 2021 IEEE/CVF Winter Conference on Applications
  of Computer Vision},
pp. \bfpage{2083}--\blpage{2092}
(\byear{2021})
\end{bchapter}
\endbibitem

%%% 15
\bibitem[\protect\citeauthoryear{Huang et~al.}{2021}]{huang2021cross}
\begin{bchapter}
\bauthor{\bsnm{Huang}, \binits{J.}},
\bauthor{\bsnm{Liu}, \binits{Y.}},
\bauthor{\bsnm{Gong}, \binits{S.}},
\bauthor{\bsnm{Jin}, \binits{H.}}:
\bctitle{Cross-sentence temporal and semantic relations in video activity
  localisation}.
In: \bbtitle{Proceedings of the 18th IEEE/CVF International Conference on
  Computer Vision},
pp. \bfpage{7199}--\blpage{7208}
(\byear{2021})
\end{bchapter}
\endbibitem

%%% 16
\bibitem[\protect\citeauthoryear{Yang et~al.}{2021}]{yang2021local}
\begin{barticle}
\bauthor{\bsnm{Yang}, \binits{W.}},
\bauthor{\bsnm{Zhang}, \binits{T.}},
\bauthor{\bsnm{Zhang}, \binits{Y.}},
\bauthor{\bsnm{Wu}, \binits{F.}}:
\batitle{Local correspondence network for weakly supervised temporal sentence
  grounding}.
\bjtitle{IEEE Transactions on Image Processing}
\bvolume{30},
\bfpage{3252}--\blpage{3262}
(\byear{2021})
\end{barticle}
\endbibitem

%%% 17
\bibitem[\protect\citeauthoryear{Duan et~al.}{2018}]{duan2018weakly}
\begin{barticle}
\bauthor{\bsnm{Duan}, \binits{X.}},
\bauthor{\bsnm{Huang}, \binits{W.}},
\bauthor{\bsnm{Gan}, \binits{C.}},
\bauthor{\bsnm{Wang}, \binits{J.}},
\bauthor{\bsnm{Zhu}, \binits{W.}},
\bauthor{\bsnm{Huang}, \binits{J.}}:
\batitle{Weakly supervised dense event captioning in videos}.
\bjtitle{Advances in Neural Information Processing Systems}
\bvolume{31},
\bfpage{1}--\blpage{11}
(\byear{2018})
\end{barticle}
\endbibitem

%%% 18
\bibitem[\protect\citeauthoryear{Lin et~al.}{2020}]{lin2020weakly}
\begin{bchapter}
\bauthor{\bsnm{Lin}, \binits{Z.}},
\bauthor{\bsnm{Zhao}, \binits{Z.}},
\bauthor{\bsnm{Zhang}, \binits{Z.}},
\bauthor{\bsnm{Wang}, \binits{Q.}},
\bauthor{\bsnm{Liu}, \binits{H.}}:
\bctitle{Weakly-supervised video moment retrieval via semantic completion
  network}.
In: \bbtitle{Proceedings of the AAAI Conference on Artificial Intelligence},
pp. \bfpage{11539}--\blpage{11546}
(\byear{2020})
\end{bchapter}
\endbibitem

%%% 19
\bibitem[\protect\citeauthoryear{Chen and Jiang}{2021}]{chen2021towards}
\begin{bchapter}
\bauthor{\bsnm{Chen}, \binits{S.}},
\bauthor{\bsnm{Jiang}, \binits{Y.-G.}}:
\bctitle{Towards bridging event captioner and sentence localizer for weakly
  supervised dense event captioning}.
In: \bbtitle{Proceedings of the 2021 IEEE/CVF Conference on Computer Vision and
  Pattern Recognition},
pp. \bfpage{8425}--\blpage{8435}
(\byear{2021})
\end{bchapter}
\endbibitem

%%% 20
\bibitem[\protect\citeauthoryear{Zheng et~al.}{2022}]{zheng2022weakly}
\begin{bchapter}
\bauthor{\bsnm{Zheng}, \binits{M.}},
\bauthor{\bsnm{Huang}, \binits{Y.}},
\bauthor{\bsnm{Chen}, \binits{Q.}},
\bauthor{\bsnm{Liu}, \binits{Y.}}:
\bctitle{Weakly supervised video moment localization with contrastive negative
  sample mining}.
In: \bbtitle{Proceedings of the AAAI Conference on Artificial Intelligence},
pp. \bfpage{3517}--\blpage{3525}
(\byear{2022})
\end{bchapter}
\endbibitem

%%% 21
\bibitem[\protect\citeauthoryear{Zhang et~al.}{2023}]{zhang2023temporal}
\begin{barticle}
\bauthor{\bsnm{Zhang}, \binits{H.}},
\bauthor{\bsnm{Sun}, \binits{A.}},
\bauthor{\bsnm{Jing}, \binits{W.}},
\bauthor{\bsnm{Zhou}, \binits{J.T.}}:
\batitle{Temporal sentence grounding in videos: A survey and future
  directions}.
\bjtitle{IEEE Transactions on Pattern Analysis and Machine Intelligence}
\bvolume{45},
\bfpage{10443}--\blpage{10465}
(\byear{2023})
\end{barticle}
\endbibitem

%%% 22
\bibitem[\protect\citeauthoryear{Liu et~al.}{2023}]{liu2023survey}
\begin{barticle}
\bauthor{\bsnm{Liu}, \binits{M.}},
\bauthor{\bsnm{Nie}, \binits{L.}},
\bauthor{\bsnm{Wang}, \binits{Y.}},
\bauthor{\bsnm{Wang}, \binits{M.}},
\bauthor{\bsnm{Rui}, \binits{Y.}}:
\batitle{A survey on video moment localization}.
\bjtitle{ACM Computing Surveys}
\bvolume{55},
\bfpage{1}--\blpage{37}
(\byear{2023})
\end{barticle}
\endbibitem

%%% 23
\bibitem[\protect\citeauthoryear{Gao et~al.}{2019}]{gao2019wslln}
\begin{botherref}
\oauthor{\bsnm{Gao}, \binits{M.}},
\oauthor{\bsnm{Davis}, \binits{L.S.}},
\oauthor{\bsnm{Socher}, \binits{R.}},
\oauthor{\bsnm{Xiong}, \binits{C.}}:
Wslln: Weakly supervised natural language localization networks.
Computing Research Repository arXiv Preprint, arXiv:1909.00239
(2019)
\end{botherref}
\endbibitem

%%% 24
\bibitem[\protect\citeauthoryear{Ma et~al.}{2020}]{ma2020vlanet}
\begin{bchapter}
\bauthor{\bsnm{Ma}, \binits{M.}},
\bauthor{\bsnm{Yoon}, \binits{S.}},
\bauthor{\bsnm{Kim}, \binits{J.}},
\bauthor{\bsnm{Lee}, \binits{Y.}},
\bauthor{\bsnm{Kang}, \binits{S.}},
\bauthor{\bsnm{Yoo}, \binits{C.D.}}:
\bctitle{Vlanet: Video-language alignment network for weakly-supervised video
  moment retrieval}.
In: \bbtitle{Computer Vision--ECCV 2020: 16th European Conference, Glasgow, UK,
  August 23--28, 2020, Proceedings, Part XXVIII},
pp. \bfpage{156}--\blpage{171}
(\byear{2020})
\end{bchapter}
\endbibitem

%%% 25
\bibitem[\protect\citeauthoryear{Chen et~al.}{2020}]{chen2020look}
\begin{botherref}
\oauthor{\bsnm{Chen}, \binits{Z.}},
\oauthor{\bsnm{Ma}, \binits{L.}},
\oauthor{\bsnm{Luo}, \binits{W.}},
\oauthor{\bsnm{Tang}, \binits{P.}},
\oauthor{\bsnm{Wong}, \binits{K.-Y.K.}}:
Look closer to ground better: Weakly-supervised temporal grounding of sentence
  in video.
Computing Research Repository arXiv Preprint, arXiv:2001.09308
(2020)
\end{botherref}
\endbibitem

%%% 26
\bibitem[\protect\citeauthoryear{Wang et~al.}{2022}]{wang2021weakly}
\begin{barticle}
\bauthor{\bsnm{Wang}, \binits{Y.}},
\bauthor{\bsnm{Deng}, \binits{J.}},
\bauthor{\bsnm{Zhou}, \binits{W.}},
\bauthor{\bsnm{Li}, \binits{H.}}:
\batitle{Weakly supervised temporal adjacent network for language grounding}.
\bjtitle{IEEE Transactions on Multimedia}
\bvolume{24},
\bfpage{3276}--\blpage{3286}
(\byear{2022})
\end{barticle}
\endbibitem

%%% 27
\bibitem[\protect\citeauthoryear{Song et~al.}{2020}]{song2020weakly}
\begin{botherref}
\oauthor{\bsnm{Song}, \binits{Y.}},
\oauthor{\bsnm{Wang}, \binits{J.}},
\oauthor{\bsnm{Ma}, \binits{L.}},
\oauthor{\bsnm{Yu}, \binits{Z.}},
\oauthor{\bsnm{Yu}, \binits{J.}}:
Weakly-supervised multi-level attentional reconstruction network for grounding
  textual queries in videos.
Computing Research Repository arXiv Preprint, arXiv:2003.07048
(2020)
\end{botherref}
\endbibitem

%%% 28
\bibitem[\protect\citeauthoryear{Zhang et~al.}{2020}]{zhang2020regularized}
\begin{bchapter}
\bauthor{\bsnm{Zhang}, \binits{Z.}},
\bauthor{\bsnm{Lin}, \binits{Z.}},
\bauthor{\bsnm{Zhao}, \binits{Z.}},
\bauthor{\bsnm{Zhu}, \binits{J.}},
\bauthor{\bsnm{He}, \binits{X.}}:
\bctitle{Regularized two-branch proposal networks for weakly-supervised moment
  retrieval in videos}.
In: \bbtitle{Proceedings of the 28th ACM International Conference on
  Multimedia},
pp. \bfpage{4098}--\blpage{4106}
(\byear{2020})
\end{bchapter}
\endbibitem

%%% 29
\bibitem[\protect\citeauthoryear{Nam et~al.}{2021}]{nam2021zero}
\begin{bchapter}
\bauthor{\bsnm{Nam}, \binits{J.}},
\bauthor{\bsnm{Ahn}, \binits{D.}},
\bauthor{\bsnm{Kang}, \binits{D.}},
\bauthor{\bsnm{Ha}, \binits{S.J.}},
\bauthor{\bsnm{Choi}, \binits{J.}}:
\bctitle{Zero-shot natural language video localization}.
In: \bbtitle{Proceedings of the 18th IEEE/CVF International Conference on
  Computer Vision},
pp. \bfpage{1470}--\blpage{1479}
(\byear{2021})
\end{bchapter}
\endbibitem

%%% 30
\bibitem[\protect\citeauthoryear{Gao and Xu}{2021}]{gao2021learning}
\begin{barticle}
\bauthor{\bsnm{Gao}, \binits{J.}},
\bauthor{\bsnm{Xu}, \binits{C.}}:
\batitle{Learning video moment retrieval without a single annotated video}.
\bjtitle{IEEE Transactions on Circuits and Systems for Video Technology}
\bvolume{32},
\bfpage{1646}--\blpage{1657}
(\byear{2021})
\end{barticle}
\endbibitem

%%% 31
\bibitem[\protect\citeauthoryear{Pennington et~al.}{2014}]{pennington2014glove}
\begin{bchapter}
\bauthor{\bsnm{Pennington}, \binits{J.}},
\bauthor{\bsnm{Socher}, \binits{R.}},
\bauthor{\bsnm{Manning}, \binits{C.D.}}:
\bctitle{Glove: Global vectors for word representation}.
In: \bbtitle{Proceedings of the 2014 Conference on Empirical Methods in Natural
  Language Processing},
pp. \bfpage{1532}--\blpage{1543}
(\byear{2014})
\end{bchapter}
\endbibitem

%%% 32
\bibitem[\protect\citeauthoryear{Radford et~al.}{2021}]{radford2021learning}
\begin{bchapter}
\bauthor{\bsnm{Radford}, \binits{A.}},
\bauthor{\bsnm{Kim}, \binits{J.W.}},
\bauthor{\bsnm{Hallacy}, \binits{C.}},
\bauthor{\bsnm{Ramesh}, \binits{A.}},
\bauthor{\bsnm{Goh}, \binits{G.}},
\bauthor{\bsnm{Agarwal}, \binits{S.}},
\bauthor{\bsnm{Sastry}, \binits{G.}},
\bauthor{\bsnm{Askell}, \binits{A.}},
\bauthor{\bsnm{Mishkin}, \binits{P.}},
\bauthor{\bsnm{Clark}, \binits{J.}},
\bauthor{\bsnm{Krueger}, \binits{G.}},
\bauthor{\bsnm{Sutskever}, \binits{I.}}:
\bctitle{Learning transferable visual models from natural language
  supervision}.
In: \bbtitle{Proceedings of the 38th International Conference on Machine
  Learning},
pp. \bfpage{8748}--\blpage{8763}
(\byear{2021})
\end{bchapter}
\endbibitem

%%% 33
\bibitem[\protect\citeauthoryear{Carreira and
  Zisserman}{2017}]{carreira2017quo}
\begin{bchapter}
\bauthor{\bsnm{Carreira}, \binits{J.}},
\bauthor{\bsnm{Zisserman}, \binits{A.}}:
\bctitle{Quo vadis, action recognition? a new model and the kinetics dataset}.
In: \bbtitle{Proceedings of the 2017 IEEE Conference on Computer Vision and
  Pattern Recognition},
pp. \bfpage{6299}--\blpage{6308}
(\byear{2017})
\end{bchapter}
\endbibitem

%%% 34
\bibitem[\protect\citeauthoryear{Vaswani et~al.}{2017}]{vaswani2017attention}
\begin{barticle}
\bauthor{\bsnm{Vaswani}, \binits{A.}},
\bauthor{\bsnm{Shazeer}, \binits{N.}},
\bauthor{\bsnm{Parmar}, \binits{N.}},
\bauthor{\bsnm{Uszkoreit}, \binits{J.}},
\bauthor{\bsnm{Jones}, \binits{L.}},
\bauthor{\bsnm{Gomez}, \binits{A.N.}},
\bauthor{\bsnm{Kaiser}, \binits{{\L}.}},
\bauthor{\bsnm{Polosukhin}, \binits{I.}}:
\batitle{Attention is all you need}.
\bjtitle{Advances in Neural Information Processing Systems}
\bvolume{30},
\bfpage{1}--\blpage{11}
(\byear{2017})
\end{barticle}
\endbibitem

%%% 35
\bibitem[\protect\citeauthoryear{Bahdanau et~al.}{2015}]{bahdanau2015neural}
\begin{bchapter}
\bauthor{\bsnm{Bahdanau}, \binits{D.}},
\bauthor{\bsnm{Cho}, \binits{K.H.}},
\bauthor{\bsnm{Bengio}, \binits{Y.}}:
\bctitle{Neural machine translation by jointly learning to align and
  translate}.
In: \bbtitle{Proceedings of the 3rd International Conference on Learning
  Representations},
pp. \bfpage{1}--\blpage{10}
(\byear{2015})
\end{bchapter}
\endbibitem

%%% 36
\bibitem[\protect\citeauthoryear{Zhou}{2009}]{zhou2009ensemble}
\begin{botherref}
\oauthor{\bsnm{Zhou}, \binits{Z.-H.}}:
Ensemble learning.
Encyclopedia of Biometrics,
270--273
(2009)
\end{botherref}
\endbibitem

%%% 37
\bibitem[\protect\citeauthoryear{Zheng et~al.}{2022}]{2zheng2022weakly}
\begin{bchapter}
\bauthor{\bsnm{Zheng}, \binits{M.}},
\bauthor{\bsnm{Huang}, \binits{Y.}},
\bauthor{\bsnm{Chen}, \binits{Q.}},
\bauthor{\bsnm{Peng}, \binits{Y.}},
\bauthor{\bsnm{Liu}, \binits{Y.}}:
\bctitle{Weakly supervised temporal sentence grounding with gaussian-based
  contrastive proposal learning}.
In: \bbtitle{Proceedings of the 2022 IEEE/CVF Conference on Computer Vision and
  Pattern Recognition},
pp. \bfpage{15555}--\blpage{15564}
(\byear{2022})
\end{bchapter}
\endbibitem

%%% 38
\bibitem[\protect\citeauthoryear{Krishna et~al.}{2017}]{krishna2017dense}
\begin{bchapter}
\bauthor{\bsnm{Krishna}, \binits{R.}},
\bauthor{\bsnm{Hata}, \binits{K.}},
\bauthor{\bsnm{Ren}, \binits{F.}},
\bauthor{\bsnm{Fei-Fei}, \binits{L.}},
\bauthor{\bsnm{Carlos~Niebles}, \binits{J.}}:
\bctitle{Dense-captioning events in videos}.
In: \bbtitle{Proceedings of the 16th IEEE International Conference on Computer
  Vision},
pp. \bfpage{706}--\blpage{715}
(\byear{2017})
\end{bchapter}
\endbibitem

%%% 39
\bibitem[\protect\citeauthoryear{Kingma and Ba}{2014}]{kingma2014adam}
\begin{botherref}
\oauthor{\bsnm{Kingma}, \binits{D.P.}},
\oauthor{\bsnm{Ba}, \binits{J.}}:
Adam: A method for stochastic optimization.
Computing Research Repository arXiv Preprint, arXiv:1412.6980
(2014)
\end{botherref}
\endbibitem

%%% 40
\bibitem[\protect\citeauthoryear{Wu et~al.}{2023}]{wu2023atomic}
\begin{bchapter}
\bauthor{\bsnm{Wu}, \binits{H.}},
\bauthor{\bsnm{Lyu}, \binits{Y.}},
\bauthor{\bsnm{Shen}, \binits{X.}},
\bauthor{\bsnm{Zhao}, \binits{X.}},
\bauthor{\bsnm{Wang}, \binits{M.}},
\bauthor{\bsnm{Zhang}, \binits{X.}},
\bauthor{\bsnm{Luo}, \binits{Z.}}:
\bctitle{Atomic-action-based contrastive network for weakly supervised temporal
  language grounding}.
In: \bbtitle{2023 IEEE International Conference on Multimedia and Expo (ICME)},
pp. \bfpage{1523}--\blpage{1528}
(\byear{2023})
\end{bchapter}
\endbibitem

%%% 41
\bibitem[\protect\citeauthoryear{Song et~al.}{2023}]{song2023marn}
\begin{barticle}
\bauthor{\bsnm{Song}, \binits{Y.}},
\bauthor{\bsnm{Wang}, \binits{J.}},
\bauthor{\bsnm{Ma}, \binits{L.}},
\bauthor{\bsnm{Yu}, \binits{J.}},
\bauthor{\bsnm{Liang}, \binits{J.}},
\bauthor{\bsnm{Yuan}, \binits{L.}},
\bauthor{\bsnm{Yu}, \binits{Z.}}:
\batitle{M\uppercase{ARN}: Multi-level attentional reconstruction networks for
  weakly supervised video temporal grounding}.
\bjtitle{Neurocomputing}
\bvolume{554},
\bfpage{126625}
(\byear{2023})
\end{barticle}
\endbibitem

\end{thebibliography}

\end{document}